\documentclass[10pt,journal,compsoc]{IEEEtran}
\ifCLASSOPTIONcompsoc
  \usepackage[nocompress]{cite}
\else
  \usepackage{cite}
\fi
\ifCLASSINFOpdf
\else
\fi
\usepackage{times}
\usepackage{epsfig}
\usepackage{bbm}
\usepackage{graphicx}
\usepackage{amsmath}
\usepackage{amssymb}
\usepackage{comment}
\usepackage{gensymb}
\usepackage{booktabs, multicol, multirow}
\usepackage{array}
\usepackage{color}
\usepackage{subfigure}
\usepackage{caption}
\usepackage{cite}
\usepackage{balance}
\usepackage{setspace}

\begin{document}
%
\title{Profile to Frontal Face Recognition in the Wild Using Coupled Conditional GAN}
%
%
%
%

\author{Fariborz Taherkhani*\thanks{* Authors Contributed Equally},~\IEEEmembership{Student Member,~IEEE,}
        Veeru Talreja* ,~\IEEEmembership{Student Member,~IEEE,}
        \\ Jeremy  Dawson,~\IEEEmembership{Member, ~IEEE,} Matthew C. 
        Valenti,~\IEEEmembership{Fellow,~IEEE,} \\
        and~ Nasser M. Nasrabadi,~\IEEEmembership{Fellow,~IEEE}}
\markboth{JOURNAL OF LATEX CLASS FILES, VOL. 14, NO. 8, AUGUST 2015}%
{Shell \MakeLowercase{\textit{et al.}}: Bare Demo of IEEEtran.cls for Biometrics Council Journals}
\IEEEtitleabstractindextext{%
\begin{abstract}
In recent years, with the advent of deep-learning, face recognition has achieved exceptional success. However, many of these deep face recognition models perform much better in handling frontal faces compared to profile faces. The major reason for poor performance in handling of profile faces is  that it is inherently difficult to learn pose-invariant deep representations that are useful for profile face recognition. In this paper, we hypothesize that the profile face domain possesses a latent connection with the frontal face domain in a latent feature subspace. We look to exploit this latent connection by projecting the profile faces and frontal faces into a common latent subspace and perform verification or retrieval in the latent domain. We leverage a coupled conditional generative adversarial network (cpGAN) structure to find the hidden relationship between the profile and frontal images in a latent common embedding subspace. Specifically, the  cpGAN framework consists of two conditional GAN-based sub-networks, one dedicated to the frontal domain and the other dedicated to the profile domain. Each sub-network tends to find a projection that maximizes the pair-wise correlation between the two feature domains in a common embedding feature subspace.  The efficacy of our approach compared with the state-of-the-art is demonstrated using the CFP, CMU Multi-PIE,  IJB-A, and IJB-C datasets. Additionally, we have also implemented a coupled convolutional neural network (cpCNN) and an adversarial discriminative domain adaptation network (ADDA) for profile to frontal face recognition. We have evaluated the performance of cpCNN and ADDA and compared it with the proposed cpGAN. Finally, we have also evaluated our cpGAN for reconstruction of frontal faces from input profile faces contained in the VGGFace2 dataset. 
\end{abstract}}

\maketitle
 \begingroup\renewcommand\thefootnote{\textsection}
\footnotetext{* Fariborz Taherkhani and Veeru Talreja contributed equally}
\endgroup
\section{Introduction}
\label{sec:introduction}
Due to the emergence of deep-learning, face recognition has achieved exceptional success in recent years \cite{Cao2018PoseRobustFR}. However, many of these deep face recognition models perform relatively poorly in handling profile faces compared to frontal faces \cite{sengupta_wacv_cfp}. When faces are captured in an unconstrained environment, in the wild, they are often in a profile orientation. Thus there is an equivalency between the challenging problems of unconstrained face recognition and profile face recognition. Pose, expression, and lighting variations are considered to be major obstacles in attaining high unconstrained face recognition performance.  Some methods \cite{Masi_pose_aware_2016,Cao2018PoseRobustFR} attempt to address the pose-variation issue by learning pose-invariant features, while some other methods \cite{Yim2015RotatingYF, Yin2017TowardsLF,Cole2017SynthesizingNF,FNM, Tran_disentangled_cvpr_17} try to normalize images while preserving identity to a single frontal pose before recognition.  However, there are three major difficulties related to face frontalization or normalization in unconstrained environments: 
\begin{itemize}
    \item Complicated face-variations besides pose: In comparison to a controlled environment, there are more complex face variations, e.g., lighting, head pose, expression, in real-world scenarios. It is a difficult task to directly
warp the input face to a normalized view \cite{FNM}. 
   \item Unpaired data: Undoubtedly, obtaining a strictly normalized face is expensive and time-consuming, but getting an effective pair of images consisting of a target normalized face (i.e., frontal-facing, neutral expression) and an input face is  difficult due to highly imbalanced datasets\cite{FNM}.  
   \item Presence of artifacts:  Synthesized ‘frontal’ faces contain artifacts caused by occlusions and non-rigid expressions.
\end{itemize} 

In this paper, we hypothesize that the profile face domain shares a latent connection with the frontal face domain in a latent deep feature subspace. We aim to exploit this connection by projecting the profile faces and frontal faces into a common latent subspace and perform verification or retrieval in this latent domain. We propose an embedding model for profile to frontal face verification based on a deep coupled learning framework which uses a generative adversarial network (GAN) to find the hidden relationship between the profile face features and frontal face features in a latent common embedding subspace. 

Our work is conceptually related to the embedding category of super-resolution \cite{shekhar2017synthesis,zhang2015coupled,jiang2016cdmma,li2019low} in that our approach also performs verification of profile and frontal faces in the latent space but not in the original image space. From our experiments, we observe that transforming profile and frontal face features into a latent embedding subspace could yield higher performance than image-level face frontalization, which is susceptible to the negative influence of artifacts as a result of image synthesis. To our best knowledge, this study is the first attempt to perform profile-to-frontal face verification in a latent embedding subspace using generative modeling. This work is an extension of our previous work \cite{taherkhani2020pf}.

This paper makes the following contributions:

\begin{itemize}
    \item The paper develops of a novel profile to frontal face recognition model using a coupled conditional GAN (cpGAN) framework with multiple loss functions.
    \item The paper includes comprehensive experiments using different datasets and a comparison of the proposed method with the state-of-the-art methods, indicating the efficacy of the proposed GAN framework.
    \item The proposed framework can potentially be used to improve the performance of traditional face recognition methods by integrating it as a preprocessing procedure for a face-frontalization schema.
    \item The paper includes experiments to evaluate the frontalization performance of the cpGAN by using a face matcher (verifier) to compare off-pose faces with a gallery of frontal faces and also compare the frontalized images with the gallery to see if frontalizing the face would increase the face matcher performance.
    \item The paper implements a coupled CNN (cpCNN) and includes experiments to evaluate the benefits of using the GAN by comparing the performance of a cpCNN with our proposed approach (cpGAN).
    \item The paper implements an adversarial discriminative domain adaptation (ADDA) framework for profile to frontal face recognition and includes experiments to compare the performance of our proposed cpGAN with an ADDA network.
   \item The paper includes generated qualitative results for the VGGFace2 dataset to test the robustness and reconstruction ability of our proposed coupled GAN framework.
    
\end{itemize}

\section{Related Work}
 \subsection{\textbf{Face recognition using Deep-Learning}}
 
 Before the advent of deep-learning, traditional methods for face recognition (FR) used one or more layer representations, such as the histogram of the feature codes, filtering responses, or distribution of the dictionary atoms \cite{Wang2018DeepFR}.  FR research was concentrated more toward separately improving preprocessing, local descriptors, and feature transformation; however, overall improvement in FR accuracy was very slow. This all changed with the advent of deep-learning, and now, deep-learning is the prominent technique used for FR. 
 
 Recently, various deep-learning models such as \cite{taherkhani2018deep, Dabouei_2020_WACV} have been used as baseline models for FR. Simultaneously, various loss functions have been explored and used in FR. These loss functions can be categorized as the Euclidean-distance-based
loss, angular/cosine-margin-based loss, and softmax loss and its variations. The contrastive loss and the triplet loss are the commonly used Euclidean-distance-based loss functions \cite{Sun_deep_learning_face_representation_nips,schroff2015facenet,taherkhani2019weakly,taherkhani2019matrix,talreja_TIFS,taherkhani_TBIOM}. For avoiding misclassification of difficult samples \cite{talreja2017multibiometric,talreja2018biometrics}, the learned face features need to be well separated. Angular/cosine-margin based loss \cite{Liu2017SphereFaceDH,Deng_2019_CVPR,mohsenvand2020contrastive} is commonly used to make the learned features more separable with a larger angular/cosine distance. Finally, in the category of softmax loss and its variants for  FR \cite{Hasnat2017vonMM,Wang_NormFace_ICM,Liu2017RethinkingFD}, the softmax loss is modified to improve the FR performance as in \cite{Liu2017RethinkingFD}, where the cosine distance among data features is optimized along with normalization of features and weights.

In the SphereFace method \cite{Liu2017SphereFaceDH}, angular discriminative features are learned using CNNs by using an  angular softmax (A-Softmax) loss. The notion behind using A-Softmax loss is that, geometrically, it can be viewed as imposing discriminative constraints on a hypersphere manifold. Recently, in order to maximize face class separability, a prominent line of research is to integrate margins in well-established loss functions. For example, in ArcFace approach \cite{Deng_2019_CVPR}, an Additive Angular Margin Loss is proposed to obtain highly discriminative features for face recognition. The ArcFace has a clear geometric interpretation due to its exact correspondence to geodesic distance on a hypersphere \cite{Deng_2019_CVPR}. In the UniformFace method \cite{duan2019uniformface}, a new supervised objective function named Uniform loss has been proposed to learn deep equidistributed representations for face recognition, where the complete feature space on the hypersphere manifold has been exploited by uniformly spreading the class centers on the manifold. A survey of deep-learning methods for face recognition can be found in \cite{guo2019survey}. 

\subsection{Generative Adversarial Networks}
Introduced by Goodfellow \textit{et al.} \cite{NIPS_GAN}, the Generative Adversarial Network (GAN) learns a generator network, G, and a discriminator network, D, with a minimax optimization procedure. Using this minimax optimization over a generator and a discriminator provides a simple yet powerful way to map from a source data distribution to a target distribution. GANs have been used for a wide range of applications such as image generation \cite{denton2015deep,dosovitskiy2015learning,taigman2016unsupervised}, 3D object generation \cite{wu2016learning}, etc. In addition to the original GAN, there have been other flavors of GAN \cite{arjovsky2017wasserstein,radford2015unsupervised,mirza_2014_conditional} that have been developed to resolve some of the issues with the original GAN.  The Wasserstein GAN \cite{arjovsky2017wasserstein} proposed the use of Wassertein distance in order to provide a more stable training of GANs.
Deep Convolutional GAN (DC-GAN) \cite{radford2015unsupervised} was an extension of the original GAN, where the multi-layer perceptron structure is replaced by convolutional structures.  Another popular extension of GAN is the Conditional GAN, which was introduced by Mirza and Osindero in \cite{mirza_2014_conditional}. In Conditional GAN, both the generator and  discriminator are conditioned on an additional variable, $x$. This additional variable could be any kind of auxiliary information such as discrete labels \cite{mirza_2014_conditional} or text \cite{RAYLLS_16}. The most recent GAN models achieve better synthesis by utilizing these conditional settings and introducing latent factors to disentangle the objective space. For instance, Info-GAN \cite{chen2016infogan} employs the latent code for information loss to regularize the generative network. There have also been many instance of GAN usage for face frontalization or generating pose invariant features. Yin \textit{et al.} \cite{yin2017towards} integrated 3D Morphable Model (3DMM) into the GAN structure to propose 3DMM conditioned Face Frontalization Generative Adversarial Network (GAN), termed as FF-GAN. Tran \textit{et al.} \cite{Tran_disentangled_cvpr_17} combined face frontalization and learning a pose-invariant representation from a non-frontal face image and integrated it with a GAN structure to propose a Disentangled Representation learning-Generative Adversarial
Network (DR-GAN). 

\subsection{\textbf{Profile-Frontal Face Recognition}}
Face recognition with pose variation in an unconstrained environment is a very challenging problem. Existing methods focus on the pose variation problem by training separate models for learning pose-invariant features \cite{Masi_pose_aware_2016, Cao2018PoseRobustFR}, elaborate
dense 3D facial landmark detection and warping \cite{taigman_deepface}, and  synthesizing a frontal, neutral expression face from a single image \cite{Tran_disentangled_cvpr_17,Yim2015RotatingYF, Yin2017TowardsLF,Cole2017SynthesizingNF,FNM}. 

\textbf{Pose-Invariant Feature Representation}: Face frontalization may be considered as an image-level pose invariant representation. However, feature-level pose invariant representations have also been a mainstay for face recognition. Canonical Correlation Analysis (CCA) was used in earlier works to analyze the commonality among pose-variant samples. Recently, with the advent of deep-learning, deep-learning-based methods have become popular for pose invariant feature representation. Cao \textit{et al.} \cite{Cao2018PoseRobustFR} exploit the inherent mapping between profile and frontal faces and transform a deep profile face representation to a canonical pose by adaptively adding residuals. Additionally, deep-learning methods consider several aspects, such as multiview perception layers \cite{zhu_MV_perceptron}, to learn a model separating identity from viewpoints. In \cite{zhu_MV_perceptron}, given a single 2D face image,  a deep neural net, named Multi-View Perceptron (MVP) can untangle the identity and view features, and infer a full spectrum of multi-view images. MVP can also predict images under viewpoints that are unobserved in the training data. To allow a single network structure for multiple pose inputs, feature pooling across different poses is proposed in \cite{Kan_multiview_deep}. There have also been methods related to pose-invariant feature disentanglement \cite{Peng_reconstr_disen} or identity preservation \cite{Yin2017MultiTaskCN,Zhu_DL_iden_preser} that aim to factorize out the non-identity part with a meticulously designed network. In \cite{Zhu_DL_iden_preser}, a new learning-based face representation, the Face Identity-Preserving (FIP) features, has been proposed. The FIP features are learned by using a deep neural network that combines the feature extraction layers and the reconstruction layer. The former layer generates FIP features from a face image, while the latter layer transforms the FIP features into an image in the canonical view.

\textbf{Face Frontalization}: Using a single image with large pose variation, it is very challenging to synthesize face with a frontal view with a neutral expression face due to two major reasons: a) recovering the 3D information from 2D projections is obscure and uncertain and b) presence of self-occlusion.   Seminal works date back
to the 3D Morphable Model (3DMM) \cite{blanz2003face}, which models
both the shape and appearance as PCA spaces. Hassner \textit{et al.} \cite{hassner_face_front} adopt a 3D shape model combined with input images to register and produce the frontalized face.  Based on 3DMM, Zhu \textit{et al.} \cite{zhu_face_norm}  provide a high-fidelity pose and expression normalization method. However, 3D-based methods often do not provide reasonable results and suffer from a significant performance drop with large pose variations due to artifacts and severe texture losses. Some deep-learning-based methods have shown promising performance in terms
of face frontalization \cite{bousmalis2017unsupervised,Cole2017SynthesizingNF,yang2015weakly,Yin2017TowardsLF,Zhao_PIM_cvpr,Tran_disentangled_cvpr_17,FNM}. In \cite{yang2015weakly}, a recurrent transform unit is
proposed to incrementally rotate faces in fixed yaw angles and synthesize discrete 3D views. FF-GAN \cite{Yin2017TowardsLF} solves the problem of large-pose face frontalization in the wild by incorporating a 3D face model into a GAN. Considering photo realistic and identity-preserving frontal view synthesis, a domain adaptation strategy for pose invariant face recognition is discussed in \cite{Zhao_PIM_cvpr}. Tran \textit{et al.} \cite{Tran_disentangled_cvpr_17} propose a GAN framework to rotate a face and disentangle the identity representation by using a given pose code. In \cite{FNM}, a face normalization model (FNM) uses a GAN network with three distinct losses for generating canonical-view and expression-free frontal images. 

\section {Generative Adversarial Network}

GAN was first introduced by Goodfellow, \textit{et al.} \cite{NIPS_GAN} and has drawn great attention from the deep-learning research community due to its remarkable performance on generative tasks. The GAN framework is based on two competing networks --- a generator network, G, and a discriminator network, D. The generator,  $G(z;\theta_{g})$,  is a differentiable function which maps the noise variable, $z$, from a training noise distribution, $p_{z}(z)$, to a data space with distribution, $p_{data}$, using the network parameters, $\theta_{g}$. On the other hand, the discriminator, $D(.;\theta_{d})$, is also a differentiable function, which discriminates between the real data, $y$, and  the generated fake data, $G(z)$, using a binary classification model. Specifically, the min-max two-player game between the generator and the discriminator provides a simple and powerful way to estimate target distribution and generate novel image samples \cite{FNM}. The loss function, $L(D,G)$, for GAN is given as:
\begin{equation}\begin{split}
     L(D,G) & = E_{y\sim P_{data}(y)}[\log D(y)]\\ & + E_{z\sim P_{z}(z)}[\log (1-D(G(z)))]. \end{split}
 \end{equation}
 The objective (two player minimax game) for GAN is as follows:
 \begin{equation}\begin{split}
     \min_{G}\max_{D} L(D,G) & =\min_{G}\max_{D}[E_{y\sim P_{data}(y)}[\log D(y)]\\ & + E_{z\sim P_{z}(z)}[\log (1-D(G(z)))]].\end{split}\label{eq:2}
 \end{equation}
 
 In conditional GAN \cite{mirza_2014_conditional}, both the generator and  discriminator are conditioned on an additional variable, $x$. The loss function for the conditional GAN is given as follows:
 
 \begin{equation}\begin{split}
     L_{c}(D,G) & = E_{y\sim P_{data}(y)}[\log D(y|x)]\\ & + E_{z\sim P_{z}(z)}[\log (1-D(G(z|x)))].\end{split}\label{eq:3}
 \end{equation} Hereafter, we will denote the objective for the conditional GAN as $F_{cGAN}(D,G,y,x)$, which is given by: \begin{equation}\begin{split}
F_{cGAN}(D,G,y,x) & = \min_{G}\max_{D} [E_{y\sim P_{data}(y)}[\log D(y|x)]\\ & + E_{z\sim P_{z}(z)}[\log (1-D(G(z|x)))]].\end{split}\label{eq:4}
 \end{equation}

\section{Proposed Method}
Here, we describe our method for profile to frontal face recognition. In contrast to the face normalization methods, we do not perform pose normalization (i.e., frontalization) on each profile  image  before matching. Instead, we seek to project the profile and frontal face images to a common latent low-dimensional embedding subspace using generative modeling. Inspired by the success of GANs \cite{NIPS_GAN}, we explore adversarial networks to project profile and frontal images to a common subspace for recognition. 
 
\begin{figure*}[t]
\centering
\captionsetup{justification=centering}
\includegraphics[width=13.5cm]{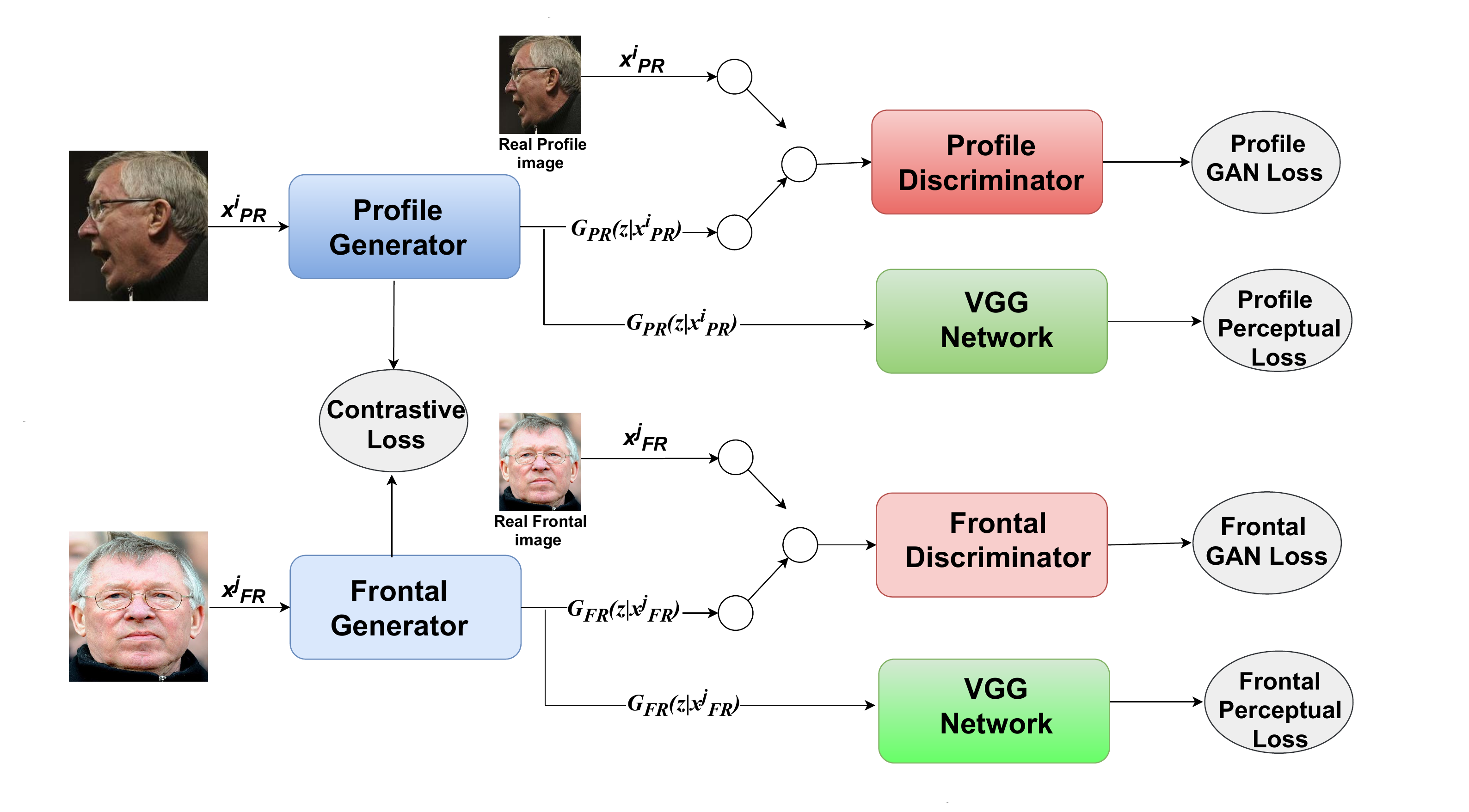}
\caption{Block diagram of PF-cpGAN.}\label{fig:arch}
\end{figure*}

 The framework of proposed profile to frontal coupled generative adversarial  network (PF-cpGAN; shown in Fig. \ref{fig:arch}) consists of two modules, where each module contains a GAN architecture comprised of a generator and a discriminator. The generators that we have used in both  modules are U-net auto-encoders that are coupled together using a contrastive loss function. In addition to  adversarial and contrastive loss, we propose to guide the generators using the perceptual loss \cite{Johnson2016PerceptualLF} based on the VGG 16 architecture, as well as an $L_2$ reconstruction error. The perceptual loss helps to generate a sharp and realistic reconstruction of the images. 
 
 \subsection{Profile to Frontal Coupled GAN}
 
 The main objective of PF-cpGAN is the recognition of profile face images with respect to a gallery of frontal face images, which have not been seen during the training. The matching of the profile and the frontal face images is performed in a common embedding subspace. PF-cpGAN consists of two modules: a profile GAN module and a frontal GAN module, both consisting of a GAN (generator + discriminator), and a perceptual network based on VGG-16.
 
\begin{figure*}[t]
\centering     
\subfigure[UNet Generator]{\label{fig:gen}\includegraphics[scale=0.7]{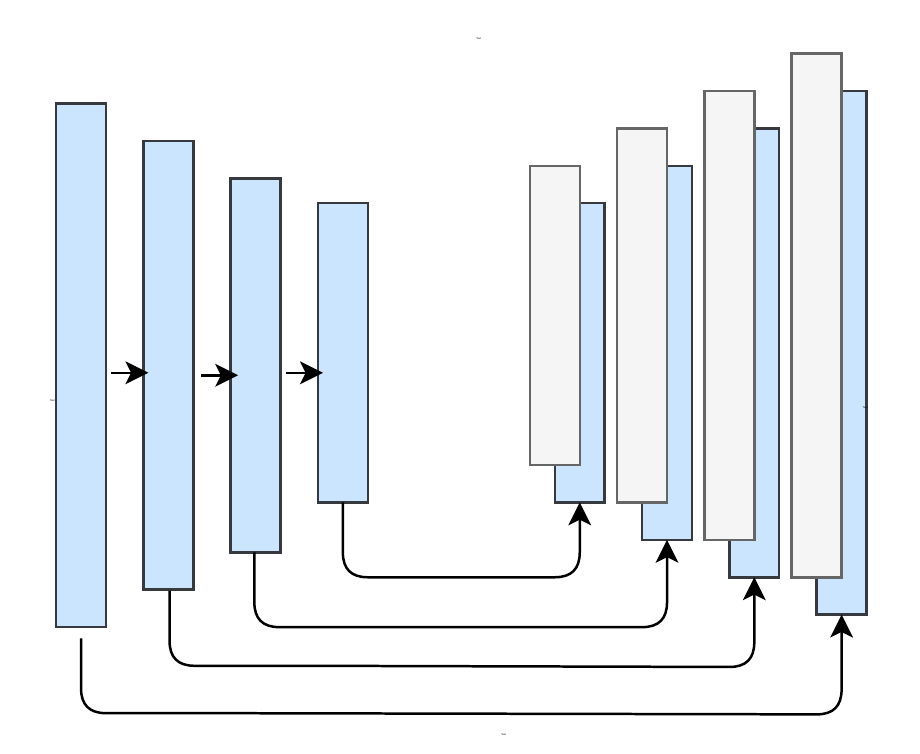}}
\subfigure[Patch-Based Discriminator]{\label{fig:disc}\includegraphics[scale=0.7]{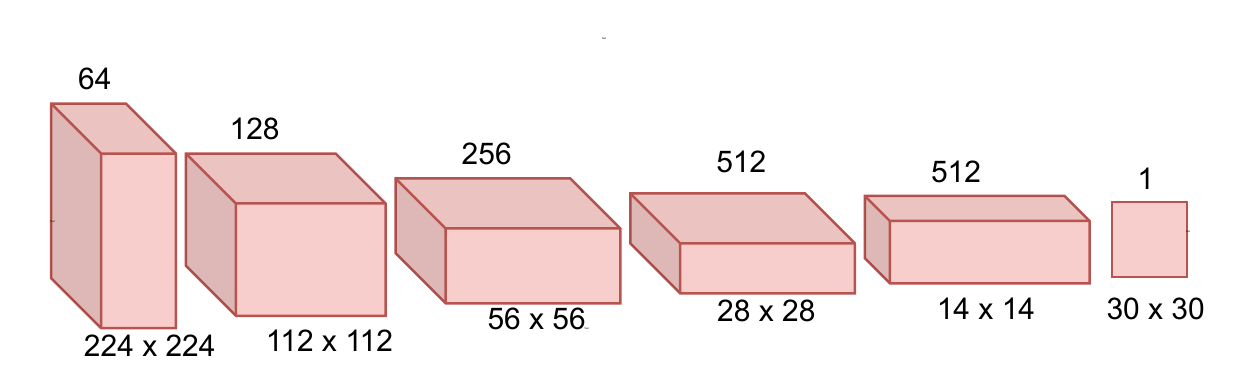}}

\caption{GAN Architectures}

\label{fig:GAN_arch}
\end{figure*}

 For the generators, we  use a U-Net \cite{ronneberger_2015_unet} auto-encoder architecture (shown in Fig. \ref{fig:gen}). The primary reason for using U-Net
is that the encoder-decoder structure tends to extract global features and generate images by leveraging this overall information, which is very useful for global shape transformation
tasks such as profile to frontal image conversion. Moreover, for many image translation problems, there is a significant amount of low-level information that needs to be shared between the input and output, and it is desirable to pass this information directly across all the layers including the bottleneck. Therefore, the use of skip-connections, as in U-net, provides a means for the encoder-decoder structure to circumvent the bottleneck and pass the information over to other layers. 

For discriminators, we have used patch-based discriminators \cite{Isola_2017_ImagetoImageTW}(shown in Fig. \ref{fig:disc}), which are trained iteratively along with the respective generators. $L_1$ loss performs very well when trying to preserve the low-frequency details but fails to preserve the high-frequency details. However, using a patch-based discriminator that penalizes structure at the scale of the patches ensures the preservation of high-frequency details, which are usually eliminated when only $L_1$ loss is used.

The final objective of  PF-cpGAN is to find the  hidden  relationship  between  the  profile face  features and frontal face features in a latent common embedding subspace. To find this common subspace between the two domains, we couple the two generators via a contrastive loss function, $L_{cont}$. 
 
 This loss function  $(L_{cont})$ is a distance-based loss function, which tries to ensure that semantically similar examples (genuine pairs, i.e., a profile image of a subject with its corresponding frontal image) are embedded closely in the common embedding subspace, and, simultaneously, semantic dissimilar examples (impostor pairs, i.e., a profile image of a subject and a frontal image of a different subject) are pushed away from each other in the common embedding subspace \cite{contrastive_2006_cvpr}. The contrastive loss function is defined as:

 \begin{equation}
\begin{split}
L_{cont}(z_1&(x^i_{PR}),z_2(x^j_{FR}),Y)= \\ & 
  (1-Y)\frac{1}{2}(D_z)^2 + (Y)\frac{1}{2}(\mbox{max}(0,m-D_z))^2,  
  \end{split}\label{eq:5}
  \end{equation}where
 $x^i_{PR}$ and $x^j_{FR}$ denote the \textit{i-th} profile and \textit{j-th} frontal face image, respectively. The variable $Y$ is a binary label, which is equal to 0 if $x^i_{PR}$ and $x^j_{FR}$ belong to the same class (i.e., genuine pair), and equal to 1 if $x^i_{PR}$ and $x^j_{FR}$ belong to a different class (i.e., impostor pair). $z_1(.)$ and $z_2(.)$ denote only the encoding functions of the U-Net auto-encoder to transform  $x^i_{PR}$ and $x^j_{FR}$, respectively into a common latent embedding subspace. The value $m$ is the contrastive margin and is used to ``tighten" the constraint. $D_z$ denotes the Euclidean distance between the outputs of the functions $z_1(x^i_{PR})$ and $z_2(x^j_{FR})$ given by:
 
 \begin{equation}
     D_z=\left\lVert z_1(x^i_{PR})-z_2(x^j_{FR})\right\rVert_2.
 \end{equation}
Therefore, if $Y=0$ (i.e., genuine pair), then the contrastive loss function $(L_{cont})$ is given as:
 \begin{equation}
L_{cont}(z_1(x^i_{PR}),z_2(x^j_{FR}),Y)  = \frac{1}{2}\left\lVert z_1(x^i_{PR})-z_2(x^j_{FR})\right\rVert^2_2, 
\end{equation} and if $Y=1$ (i.e., impostor pair), then contrastive loss function $(L_{cont})$ is :
  \begin{equation}
  \begin{split}
L_{cont}(z_1(x^i_{PR}),&z_2(x^j_{FR}),Y)  = \\ & \frac{1}{2}\mbox{max}\biggl(0,m-\left\lVert z_1(x^i_{PR})-z_2(x^j_{FR})\right\rVert_2\biggr)^2.
\end{split}
\end{equation}

 Thus, the total loss  for coupling the profile generator and the frontal generator  is denoted by $L_{cpl}$ and is given as: 



\begin{equation}
    \begin{split}
        L_{cpl}=\frac{1}{N^2}\sum_{i=1}^{N}\sum_{j=1}^{N}L_{cont}(z_1(x^i_{PR}),z_2(x^j_{FR}),Y), 
    \end{split}\label{eq:6}
\end{equation}
where N is the number of training samples. The contrastive loss in the above equation can also be replaced by some other distance-based metric, such as the Euclidean distance. However, the main aim of using the contrastive loss is to be able to use the class labels implicitly and find the discriminative embedding subspace, which may not be the case with some other metric such as the Euclidean distance. This discriminative embedding subspace would be useful for matching of a profile image against a frontal image.

\subsection{Generative Adversarial Loss}

Let the profile and frontal generators that reconstruct the corresponding profile and frontal image from the input profile  and frontal image, be denoted as $G_{PR}$ and $G_{FR}$, respectively. The patch-based discriminators used for the profile and frontal GANs are denoted as $D_{PR}$ and $D_{FR}$, respectively. For the proposed method, we have used the conditional GAN, where  the generator networks $G_{PR}$ and $G_{FR}$  are conditioned on input profile and frontal face images, respectively.  We have used the conditional GAN loss function \cite{mirza_2014_conditional} to train the generators and the corresponding discriminators in order to ensure that the discriminators cannot distinguish the  images reconstructed by the generators from the corresponding ground truth images. Let $L_{PR}$ and $L_{FR}$ denote the conditional GAN loss functions for the profile and the frontal GANs, respectively, where $L_{PR}$ and $L_{FR}$ are given as: 

\begin{equation}
L_{PR}=F_{cGAN}(D_{PR},G_{PR},y^i_{PR},x^i_{PR}),
\end{equation}
\begin{equation}
L_{FR}=F_{cGAN}(D_{FR},G_{FR},y^j_{FR},x^j_{FR}), 
\end{equation}where function $F_{cGAN}$ is the conditional GAN objective defined in (\ref{eq:4}). The term $x^i_{PR}$ denotes the profile image used as a condition for the profile GAN, and $y^i_{PR}$  denotes the real profile image.  Note that the real profile image $y^i_{PR}$ and the network condition given by $x^i_{PR}$ are the same. Similarly, $x^j_{FR}$ denotes the frontal image used as a condition for the frontal GAN and $y^j_{FR}$  denotes the real frontal image. Again, the real frontal image $y^j_{FR}$ and the network condition given by $x^j_{FR}$ are the same. The total loss for the coupled conditional GAN is given by:


\begin{equation}
L_{GAN}=L_{PR}+L_{FR}.
\end{equation}




\subsection{$L_2$ Reconstruction Loss}

We also consider the $L_2$ reconstruction loss for both the profile GAN and frontal GAN. The $L_2$ reconstruction loss measures the reconstruction error in terms of the Euclidean distance between the reconstructed image and the corresponding real image. Let $L_{2_{PR}}$ denote the reconstruction loss for the profile GAN and be defined as: 
\begin{equation}
    L_{2_{PR}}=\left\lVert G_{PR}(z|x^i_{PR})-y^i_{PR}\right\rVert^2_2,
\end{equation}where $y^i_{PR}$ is the ground truth profile image $G_{PR}(z|x^i_{PR})$ is the output of the profile generator.

Similarly, Let $L_{2_{FR}}$ denote the reconstruction loss for the frontal GAN: 

\begin{equation}
    L_{2_{FR}}=\left\lVert G_{FR}(z|x^j_{FR})-y^j_{FR}\right\rVert^2_2,
\end{equation}where $y^j_{FR}$ is the ground truth frontal image, $G_{FR}(z|x^j_{FR})$ is the output of the frontal generator.

The total $L_2$ reconstruction loss function is given by:
\begin{equation}
    L_{2}=\frac{1}{N^2}\sum_{i=1}^{N}\sum_{j=1}^{N}(L_{2_{PR}}+L_{2_{FR}}).
\end{equation}

\subsection{Perceptual Loss}\label{subsec:percloss}
In addition to the GAN loss and the reconstruction loss that are used to guide the generators, we have also used the perceptual loss, which was introduced in \cite{Johnson2016PerceptualLF} for style transfer and super-resolution. The perceptual loss function is used to compare high level differences, like content and style discrepancies, between images. The perceptual loss function involves comparing two images based on high-level representations from a pre-trained CNN, such as VGG-16 \cite{simonyan2014very}. The perceptual loss function is a good alternative to solely using  $L_1$ or $L_2$ reconstruction error, as it gives better and sharper high-quality reconstruction images \cite{Johnson2016PerceptualLF}.    

In our proposed approach, perceptual loss is added to both the profile and the frontal module using a pre-trained VGG-16 network\cite{simonyan2014very} . We extract the high-level features (ReLU3-3 layer) of the VGG-16 for both the real input image and the reconstructed output of the U-Net generator. The $L_1$ distance between these features of real and reconstructed images is used to guide the generators $G_{PR}$ and $G_{FR}$.  The perceptual loss for the profile network is defined as:

\begin{equation}
    \begin{split}
        L_{P_{PR}}=&\frac{1}{C_pW_pH_p}\sum_{c=1}^{C_{p}}\sum_{w=1}^{W_{p}}\sum_{h=1}^{H_{p}} \\ & \left\lVert V(G_{PR}(z|x^i_{PR}))^{c,w,h}-V(y^i_{PR})^{c,w,h}\right\rVert,
    \end{split}
\end{equation}
where $V(.)$ denotes a particular layer of the VGG-16, and the layer dimensions are given by $C_p$, $W_p$, and $H_p$.

Likewise the perceptual loss for the frontal network is:

\begin{equation}
    \begin{split}
        L_{P_{FR}}=&\frac{1}{C_pW_pH_p}\sum_{c=1}^{C_{p}}\sum_{w=1}^{W_{p}}\sum_{h=1}^{H_{p}} \\ & \left\lVert V(G_{FR}(z|x^j_{FR}))^{c,w,h}-V(y^j_{FR})^{c,w,h}\right\rVert.
    \end{split}
\end{equation}

The total perceptual loss function is given by:
\begin{equation}
    L_{P}=\frac{1}{N^2}\sum_{i=1}^{N}\sum_{j=1}^{N}(L_{P_{PR}}+L_{P_{FR}}).
\end{equation}

\subsection{Overall Objective Function}
The overall objective function for learning the network parameters in the proposed method is given as the sum of all the  loss functions defined above:
\begin{equation}
    \begin{split}
       L_{tot}=L_{cpl}+ \lambda_1 L_{GAN} + \lambda_2 L_{P}+ \lambda_3 L_2,
    \end{split}\label{eq:19}
\end{equation}
where $L_{cpl}$ is the coupling loss given by (9), $L_{GAN}$ is the total generative adversarial loss given by (12), $L_{P}$ is the total perceptual loss given by (18), and $L_2$ is the total reconstruction error given by (15). Variables $\lambda_1, \lambda_2,$ and $\lambda_3$ are the hyper-parameters to weigh the different loss terms. 

\section{Experiments}

 We initially describe our training setup and the datasets that we have used in our experiments. We show the efficiency of our method for the task of frontal to profile face verification by comparing its performance with state-of the-art face verification  methods  across pose-variation. We also explore the effect of face yaw in our  algorithm. Additionally, we have implemented a coupled convolutional neural network (cpCNN) and an adversarial discriminative domain adaptation network (ADDA) for profile to frontal face recognition. We have evaluated the performance of cpCNN and ADDA and compared it with the proposed PF-cpGAN. We have also evaluated our PF-cpGAN for reconstruction of frontal images from input profile images. Finally, we conduct an ablation study to investigate the effect of each term in our total training loss function in (\ref{eq:20}).

\subsection{Experimental Details}


The CMU Multi-PIE database \cite{CMU-PIE} contains 750,000 images of 337 subjects. Subjects were imaged from  15 viewing angles and 19 illumination conditions while exhibiting a range of facial expressions. It is the largest database for graded evaluation with respect to pose, illumination, and expression variations. There are four sessions in this database. For fair comparison, the database setting was made consistent with CAPG-GAN \cite{hu2018pose}, where  250 subjects from Multi-PIE have been used. Consistent with CAPG-GAN, face
images with neutral expression under 20 illuminations and
13 poses within $\pm 90\degree$ are used. We follow the setting-1 testing protocol provided in CAPG-GAN.

In setting-1, only images from session 1, which contains faces of 250 subjects were used. First 150 identities were used in the training set and remaining 100 identities were used for testing. The training set consists of all the images (13 poses and 20 illumination levels) of 150 identities, i.e., $150 \times 13 \times 20 = 39,000$ images in total. For testing, one gallery image with frontal view and normal illumination is used for each of the remaining 100 subjects. The numbers of the probe and gallery sets are 24,000 and 100 respectively


The IARPA Janus Benchmark A (IJB-A) \cite{IJB-A} is a  challenging dataset collected under complete unconstrained conditions covering full pose variation (yaw angles $-90\degree$ to $+90\degree$). IJB-A contains 500 subjects with 5,712 images and 20,414 frames extracted from videos. Following the standard protocol in \cite{IJB-A}, we evaluate our method on both verification and identification.  The IARPA Janus Benchmark C (IJB-C) dataset \cite{IJB-C} builds on IJB-A, and IJB-B \cite{IJB-B} datasets and has a total of 31,334 images for a total number of 3,531 subjects. We have also evaluated our method on IJB-A and IJB-C datasets.


VGGFace2 is a large-scale face recognition dataset, where the images are downloaded from Google Image Search and have large variations in pose, age, illumination, and ethnicity. The dataset contains about 3.3 million images corresponding to more than 9000 identities with an average of 364 images per subject.

\textbf{Implementation Details}: We have implemented a U-Net autoencoder with a ResNet-18 \cite{He2015_Resnet} architecture pre-trained on ImageNet. We have added an additional fully-connected layer after the average pooling layer for the ResNet-18 for our U-Net encoder. The U-Net decoder has the same number of layers as the encoder. The entire framework has been implemented in Pytorch. For convergence,  $\lambda_1$ is set to 1, and $\lambda_2$, and $\lambda_3$ are both set to 0.25. We used a batch size of 128 and an Adam optimizer \cite{Kingma2015AdamAM} with first-order momentum of 0.5, and learning rate of 0.0004. We have used the ReLU activation function for the generator and Leaky ReLU with a slope of 0.3 for the discriminator.  


For training,  genuine and impostor pairs were required. The genuine/impostor pairs are created by frontal and profile images of the same/different subject. During the experiments, we ensure that  the training set are balanced  by using the same number of genuine and impostor pairs. 

\subsection{Evaluation on CFP with Frontal-Profile Setting}

We first perform evaluation on the CFP dataset\cite{sengupta_wacv_cfp}, a challenging dataset created to examine the problem of frontal to profile face verification in the wild. The same 10-fold protocol is applied on both the Frontal-Profile and Frontal-Frontal settings. For fair comparison and as given in \cite{sengupta_wacv_cfp}, we consider different types of feature extraction techniques like HoG\cite{hog}, LBP\cite{lbp}, and Fisher Vector\cite{fisher_vector_dml} along with metric learning techniques like Sub-SML\cite{sub_sml}, and the Diagonal metric learning (DML) as reported in \cite{fisher_vector_dml}. We also compare against deep-learning techniques, including Deep Features\cite{deep_features}, and PR-REM\cite{Cao2018PoseRobustFR}. The results are summarized in Table \ref{table:table_cfp}. 

\begin{table}[t]
\centering

\caption{Performance comparison on CFP dataset. Mean Accuracy \\ and equal error rate (EER) with standard deviation over 10 folds.}
\scalebox{0.7}{\begin{tabular}{c c c c c}
 \hline
 &\multicolumn{2}{c}{Frontal-Profile} &\multicolumn{2}{c}{Frontal-Frontal}\\ [0.5ex] 
 \hline
 Algorithm&Accuracy&EER&Accuracy&EER  \\ \hline
 HoG+Sub-SML \cite{sengupta_wacv_cfp} &$77.31 \pm 1.61$ & $22.20 \pm 1.18 $&$88.34 \pm 1.31 $&$11.45 \pm 1.35 $ \\ \hline
  LBP+Sub-SML \cite{sengupta_wacv_cfp} &$70.02 \pm 2.14 $&$29.60 \pm 2.11 $& $83.54 \pm 2.40 $ & $16.00 \pm 1.74 $ \\ \hline
   FV+Sub-SML \cite{sengupta_wacv_cfp} &$80.63 \pm 2.12 $&$19.28 \pm 1.60 $& $91.30 \pm 0.85 $& $8.85 \pm 0.74 $ \\ \hline
    FV+DML \cite{sengupta_wacv_cfp}  &$58.47 \pm 3.51 $&$38.54 \pm 1.59 $&$91.18 \pm 1.34 $&$8.62 \pm 1.19 $ \\ \hline
    Deep Features \cite{deep_features} &$84.91 \pm 1.82 $&$14.97 \pm 1.98 $&$96.40 \pm 0.69 $&$3.48 \pm 0.67 $ \\ \hline
    PR-REM \cite{Cao2018PoseRobustFR} &$93.25 \pm 2.23 $&$7.92 \pm 0.98 $&$98.10 \pm 2.19 $&$1.10 \pm 0.22 $ \\ \hline
    PF-cpGAN&$93.78 \pm 2.46 $&$7.21 \pm 0.65 $&$98.88 \pm 1.56 $&$0.93 \pm 0.14 $ \\ \hline
\end{tabular}}
\label{table:table_cfp}
\end{table}

We can observe from Table \ref{table:table_cfp} that our proposed framework, PF-cpGAN, gives much better performance than the methods that use standard hand-crafted features of HoG, LBP, or FV, providing  minimum of $13\%$ improvement in accuracy with a $12\%$ decrease in EER for the profile-frontal setting. PF-cpGAN also improves on the performance of the Deep Features by approximately $9\%$  with a $7.5\%$ decrease in EER for the profile-frontal setting. Finally, PF-cpGAN performs on-par with the best deep-learning method of PR-REM, and, in-fact, does slightly better than  PR-REM by  $\approx 0.5\%$ improvement in accuracy with a $0.7\%$ decrease in EER for the profile-frontal setting. This performance improvement clearly shows that usage of a GAN framework for projecting the profile and frontal images in the latent embedding subspace and maintaining the semantic similarity in the latent space is better than some other deep-learning techniques such as Deep Features or PR-REM.

\subsection{Evaluation on IJB-A and IJB-C}

Here, we focus on unconstrained face recognition on the IJB-A dataset to quantify the superiority of our PF-cpGAN for profile to frontal face recognition. Some of the baselines for comparison on IJB-A are DR-GAN \cite{Tran_disentangled_cvpr_17}, FNM \cite{FNM}, PR-REM \cite{Cao2018PoseRobustFR}, and FF-GAN \cite{Yin2017TowardsLF}. We have also compared them with other methods as listed in \cite{FNM} and shown in Table \ref{table:table_ijb}. As shown in Table \ref{table:table_ijb}, we perform better than the state-of-the-art methods for both verification and identification. Specifically, for verification, we improve the genuine accept rate (GAR) by at least $1.4\%$ compared to other
methods. For instance, at the false accept rate (FAR) of 0.01, the best previously-used method is PR-REM, with an average GAR of $94.4\%$. PF-cpGAN improves upon PR-REM and gives an average GAR of $95.8\%$ at the same FAR.  We also show improvement in  identification. Specifically, the rank-1 recognition rate shows an improvement of around $1.6\%$ in comparison to the best state-of-the-art method, FNM \cite{FNM}.       

\begin{table}[h]
\centering
\caption{Performance comparison on IJB-A benchmark. Results reported are the \\ 'average$\pm$standard deviation' over the 10 folds specified in the IJB-A protocol. \\ Symbol '-' indicates that the metric is not available for that protocol.}
\scalebox{0.65}{\begin{tabular}{c c c c c}
 \hline
&\multicolumn{2}{c}{Verification} &\multicolumn{2}{c}{Identification}\\ [0.5ex]
\hline
Method &GAR@ FAR$=0.01$&GAR@ FAR$=0.001$&@ Rank-1 &@ Rank-5  \\ \hline \hline
 OPENBR \cite{klare2015pushing} & $23.6\pm0.9$&$10.4\pm1.4$&$24.6\pm1.1$&$37.5\pm0.8$\\
 GOTS \cite{klare2015pushing} &$40.6\pm1.4$&$19.8\pm0.8$&$43.3\pm2.1$&$59.5\pm2.0$ \\
 PAM \cite{Masi_pose_aware_2016} &$73.3\pm1.8$&$55.2\pm3.2$&$77.1\pm1.6$&$88.7\pm0.9$ \\
 DCNN \cite{chen2016unconstrained} &$78.7\pm4.3$&-&$85.2\pm1.8$&$93.7\pm1.0$ \\
 DR-GAN \cite{tran2017disentangled} &$77.4\pm2.7$&$53.9\pm4.3$&$85.5\pm1.5$&$94.7\pm1.1$ \\
 FF-GAN \cite{yin2017towards} &$85.2\pm1.0$&$66.3\pm3.3$&$90.2\pm0.6$&$95.4\pm0.5$ \\
 FNM \cite{FNM} &$93.4\pm0.9$&$83.8\pm2.6$&$96.0\pm0.5$&$98.6\pm0.3$ \\
 PR-REM \cite{Cao2018PoseRobustFR} &$94.4\pm0.9$&$86.8\pm1.5$&$94.6\pm1.1$&$96.8\pm1.0$ \\
 PF-cpGAN&$95.8\pm0.82$&$91.2\pm1.3$&$97.6\pm1.0$&$98.8\pm0.4$ \\
\end{tabular}}
\label{table:table_ijb}
\end{table}

\begin{table}[h]
\centering

\caption{Performance comparison on IJB-C benchmark. Results reported are the \\ 'average$\pm$standard deviation' over the 10 folds specified in the IJB-C protocol. \\ Symbol '-' indicates that the metric is not available for that protocol.}
\scalebox{0.65}{\begin{tabular}{c c c c c}
 \hline
 &\multicolumn{2}{c}{Verification} &\multicolumn{2}{c}{Identification}\\ [0.5ex] 
  \hline

 Method &GAR@ FAR$=0.01$&GAR@ FAR$=0.001$&@ Rank-1 &@ Rank-5  \\ \hline \hline
 GOTS \cite{IJB-C} &$62.1\pm1.1$&$36.3\pm1.2$&$38.5\pm1.6$&$53.8\pm1.8$ \\
 FaceNet \cite{schroff2015facenet}
 &$82.3\pm1.18$&$66.3\pm1.3$&$70.4\pm1.2$&$78.8\pm2.3$ \\
 VGG-CNN \cite{Vgg_CNN}
 &$87.2\pm1.09$&$74.3\pm0.9$&$79.6\pm1.04$&$87.8\pm1.3$ \\
 FNM \cite{FNM} &$91.2\pm0.8$&$80.4\pm1.8$&$84.6\pm0.6$&$93.7\pm0.9$ \\
 PR-REM \cite{Cao2018PoseRobustFR} &$92.1\pm0.8$&$83.4\pm1.5$&$83.1\pm0.4$&$92.6\pm1.1$\\
 PF-cpGAN&$93.8\pm0.67$&$86.1\pm0.7$&$88.3\pm1.2$&$94.8\pm0.6$ \\
\end{tabular}}
\label{table:table_ijb_c}
\end{table}

\begin{figure*}[t]
\centering     
\subfigure[IJBA]{\label{fig:a}\includegraphics[scale=0.7]{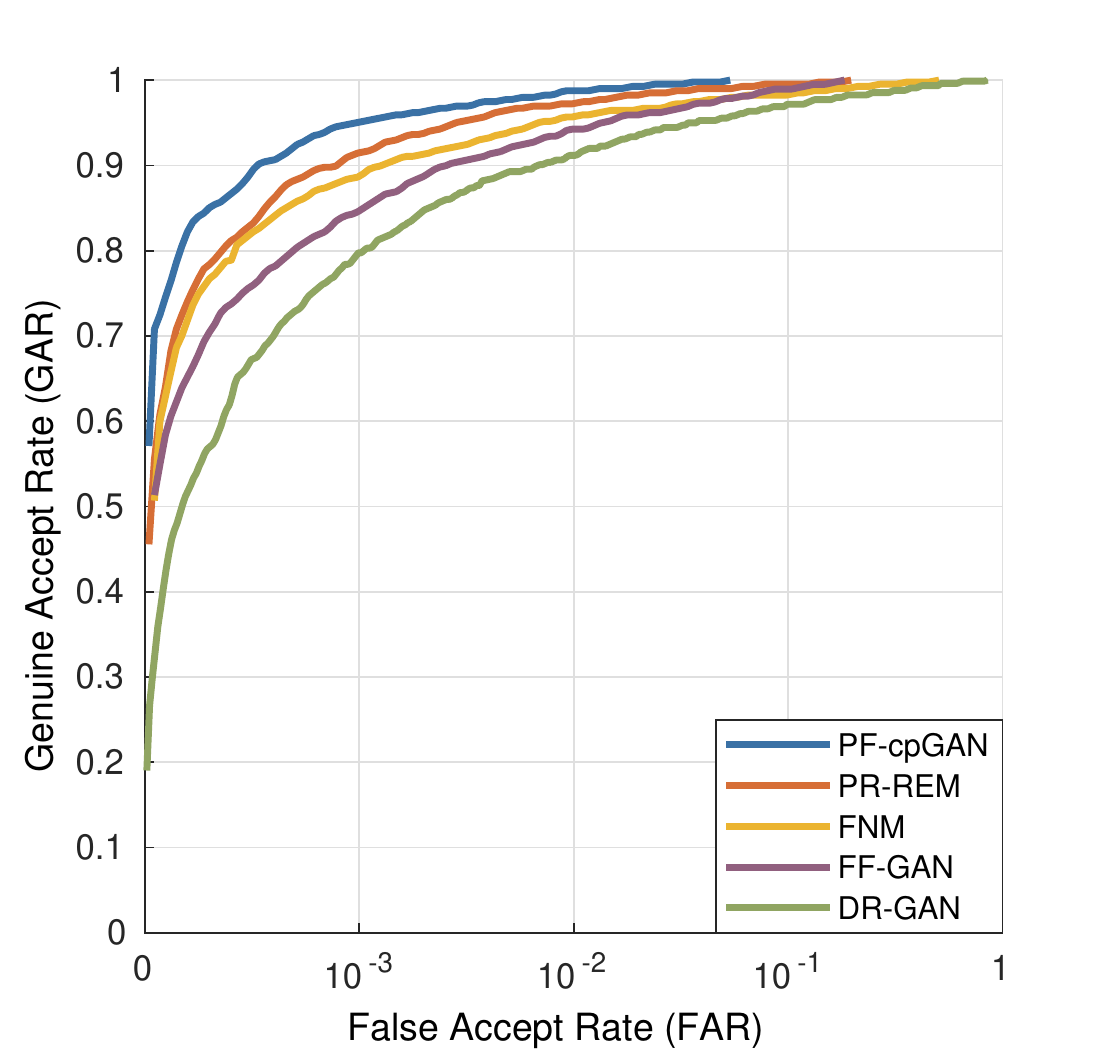}}
\subfigure[CMU Multi-PIE]{\label{fig:b}\includegraphics[scale=0.7]{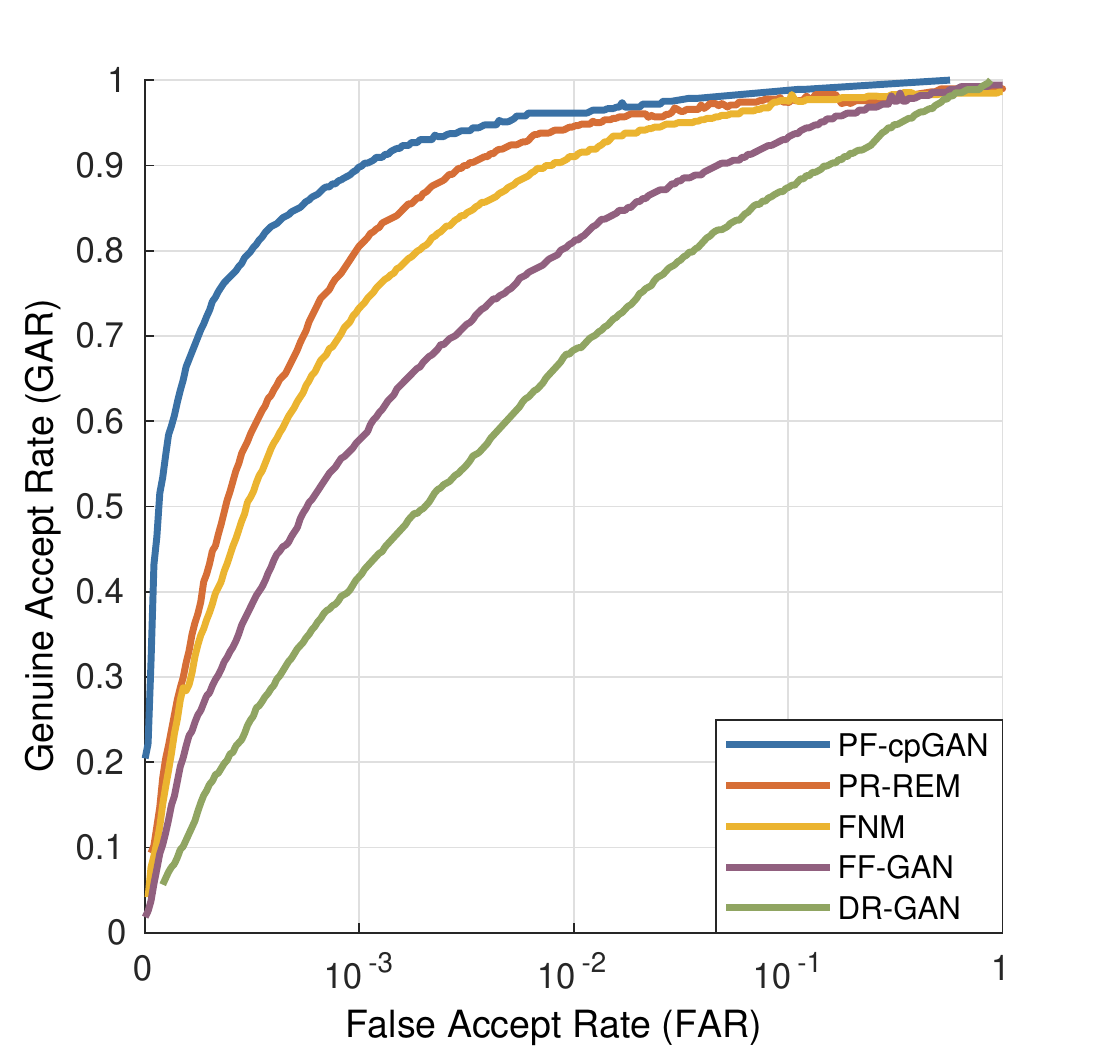}}

\caption{ROC curve comparison against the baselines for different datasets is shown in (a) and (b).}

\label{fig:ROC}
\end{figure*}

We have also plotted the receiver operating characteristic (ROC) curve and compared with the baselines given above. The ROC curves for the IJB-A dataset are given in Fig. \ref{fig:a}. As we can clearly see from the curves, the proposed  PF-cpGAN method improves upon other methods and gives much better performance, even at a FAR of 0.001.

We have also performed the task of verification and identification using the IJB-C dataset according to the verification and the identification protocol given in that dataset. The results are provided in Table \ref{table:table_ijb_c}, showing that the proposed PF-cpGAN improves on the existing state-of-the-art methods for both verification and identification. For instance, at the false accept rate (FAR) of 0.01, the best previously-used method is PR-REM, with an average GAR of $92.1\%$. PF-cpGAN improves upon PR-REM and gives an average GAR of $93.8\%$ at the same FAR. We also observe that, for  identification, specifically, rank-1 recognition, PF-cpGAN shows an improvement over the previous best state-of-the-art method FNM \cite{FNM} by about $1.1\%$.

\subsection{A Further Analysis on Influences of Face Yaw}

In addition to complete profile to frontal face recognition, we also perform a more in-depth analysis on the influence of face yaw angle on the performance of face recognition to better understand the effectiveness of  the PF-cpGAN for profile to frontal face recognition. We perform this experiment for the CMU Multi-PIE dataset \cite{CMU-PIE} under setting-1 for fair comparison with other state-of-the-art methods. As shown in Table \ref{table:multi-pie-yaw}, we achieve comparable performance with other state-of-the-art methods for different yaw angles. Under extreme pose, PF-cpGAN achieves significant
improvements (i.e., approx. $77\%$ to $88\%$ under $\pm90\degree$).  

\begin{table}[h]
\centering
\captionsetup{width=.8\linewidth}
\caption{Rank-1 recognition rates ($\%$) across poses and illuminations \\ under Multi-PIE Setting-1.}
\scalebox{0.85}{\begin{tabular}{c| c c c c c c}
 \hline
\multicolumn{1}{c}{Method} &\multicolumn{1}{c}{$\pm90\degree$} &\multicolumn{1}{c}{$\pm75\degree$}&\multicolumn{1}{c}{$\pm60\degree$}&\multicolumn{1}{c}{$\pm45\degree$}&\multicolumn{1}{c}{$\pm30\degree$}&\multicolumn{1}{c}{$\pm15\degree$}\\ [0.5ex] 
 \hline \hline
 HPN \cite{ding2017pose} & $29.82$&$47.57$&$61.24$&$72.77$&$78.26$&$84.23$\\
 c-CNN \cite{xiong2015conditional} & $47.26$&$60.7$&$74.4$&$89.0$&$94.1$&$97.0$\\
 TP-GAN \cite{huang2017beyond} & $64.0$&$84.1$&$92.9$&$98.6$&$99.99$&$99.8$\\
 PIM \cite{zhao2018towards} & $75.0$&$91.2$&$97.7$&$98.3$&$99.4$&$99.8$\\
 CAPG-GAN \cite{hu2018pose} & $77.1$&$87.4$&$93.7$&$98.3$&$99.4$&$99.99$\\
 FNM$+$VGG-Face \cite{FNM} & $41.1$&$67.3$&$83.6$&$93.6$&$97.2$&$99.0$\\
 FNM$+$Light CNN \cite{FNM}& $55.8$&$81.3$&$93.7$&$98.2$&$99.5$&$99.9$\\
 PF-cpGAN& $88.1$&$94.2$&$97.6$&$98.9$&$99.9$&$99.9$\\

\end{tabular}}
\label{table:multi-pie-yaw}
\end{table}

For further testing on the Multi-PIE dataset under setting-1, we have also plotted ROC curves and compared with other state-of-the-art methods. The ROC curves for Multi-PIE dataset are given in Fig. \ref{fig:b}.  The curves clearly indicate that the proposed method of PF-cpGAN improves upon other methods and gives much better performance, even at FAR of 0.001.

\begin{figure*}[h]
\centering
\includegraphics[scale=0.15]{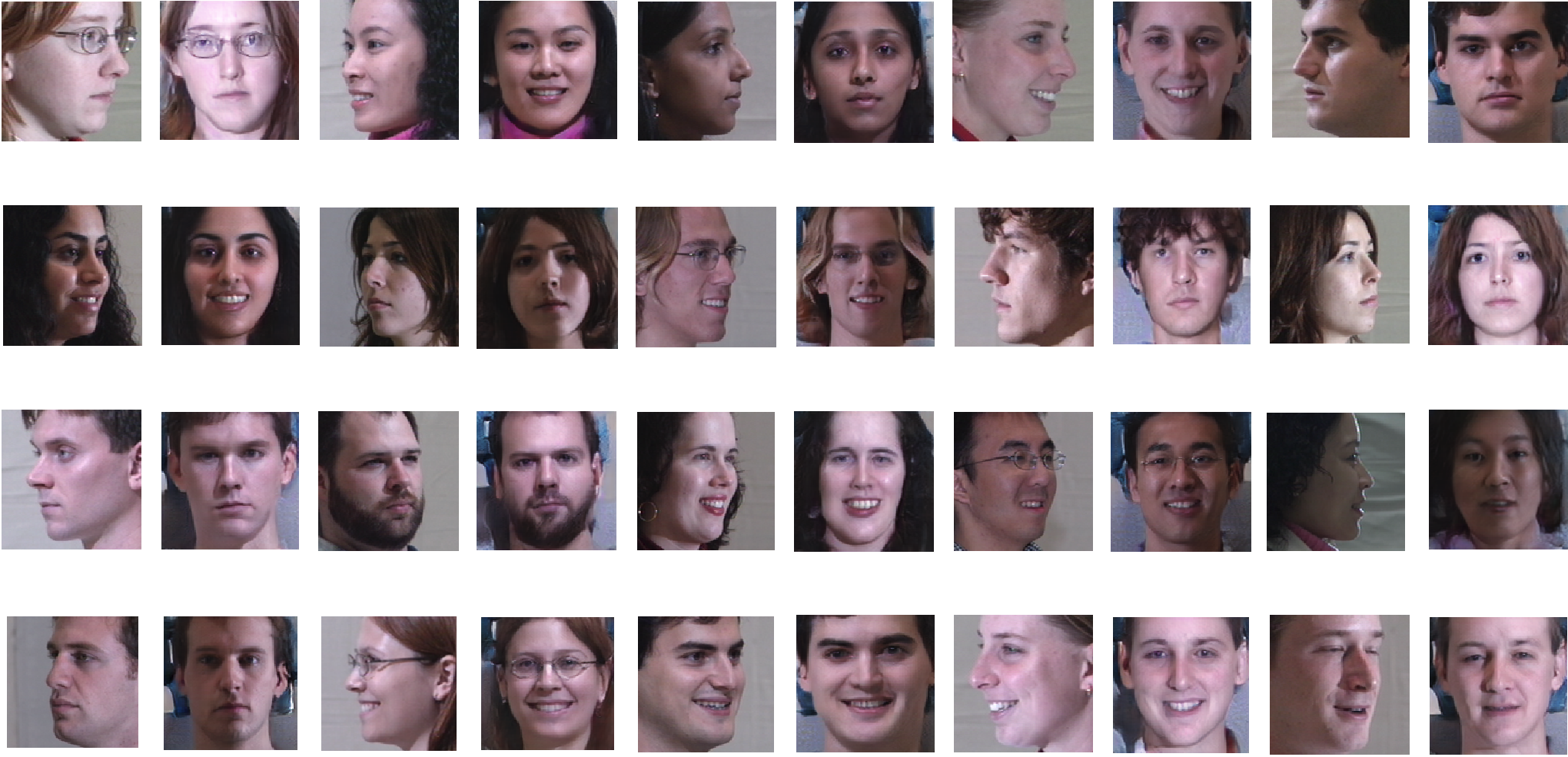}
\caption{Reconstruction of frontal images at the output of the  frontal U-Net generator with profile images as input to the profile U-Net generator. Every odd number column represent the input profile image and every even number column represents the output frontal image. The input images belong to the CMU-MultiPIE dataset. }\label{fig:profile_frontal_images}
\end{figure*}

\begin{figure*}[h]
\centering

\includegraphics[scale=0.15]{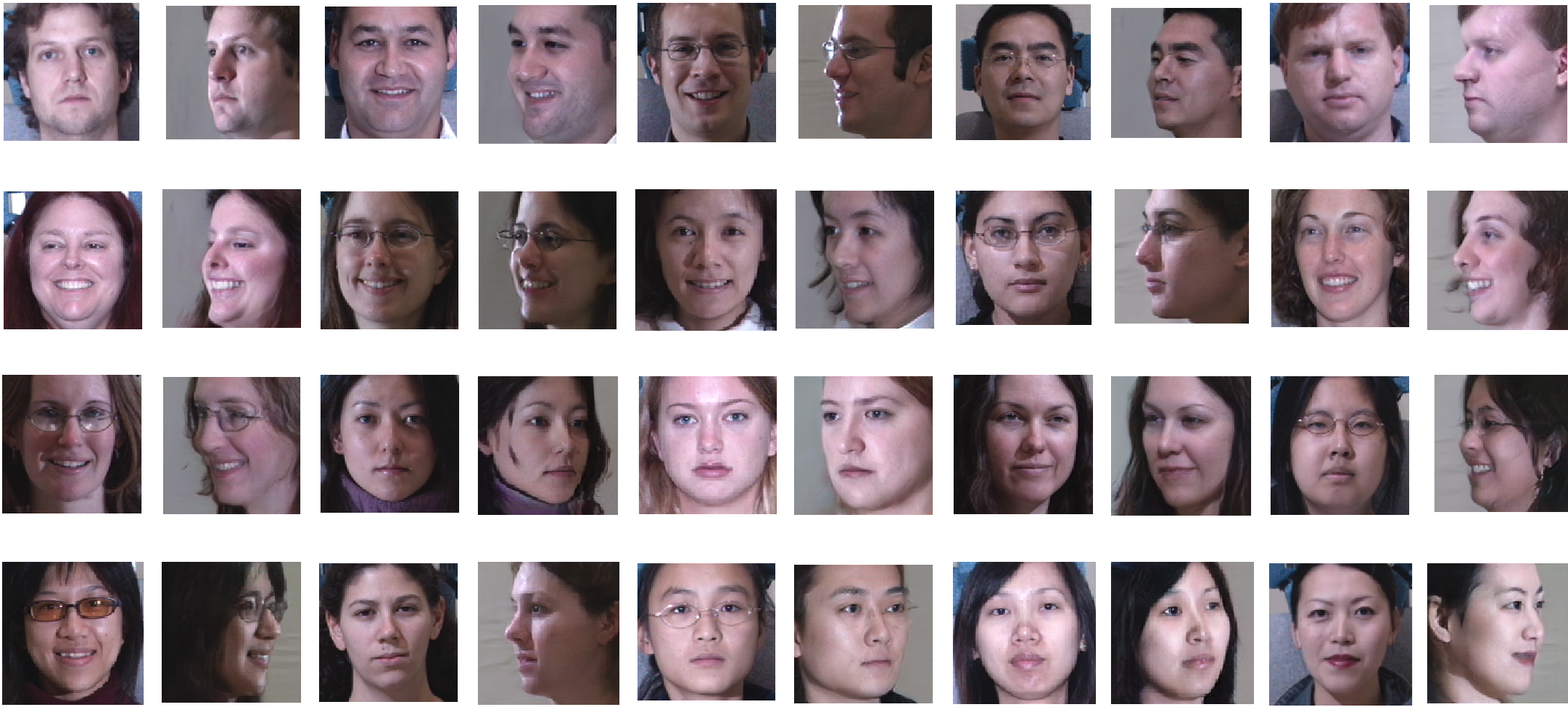}
\caption{Reconstruction of profile images at the output of the  profile U-Net generator with frontal images as input to the frontal U-Net generator. Every odd number column represents the input frontal image, and every even number column represents the output profile image. The input images belong to the CMU-MultiPIE dataset.}\label{fig:frontal_profile_images}
\end{figure*}

\subsection{Reconstruction of Frontal and Profile Images}\label{subsec:recon}

As noted in Sec. 1, the PF-cpGAN framework can also be used for reconstruction of frontal images by using profile images as input and vice versa. The results of reconstructing frontal images  using the profile images as input are given in Fig. \ref{fig:profile_frontal_images}, and the results of reconstructing  profile images  using the frontal images as input is given in Fig. \ref{fig:frontal_profile_images}. The reconstruction procedure for frontal images is given as follows: The profile image is given as input to the profile U-Net generator and the feature vector generated at the bottleneck of the profile generator (i.e., at the output of the encoder of the profile U-Net generator) is passed through the decoder section of the frontal  U-Net generator to reconstruct the frontal image. Similarly the reconstruction procedure for profile images is given as follows: The frontal image is given as input to the frontal U-Net generator, and the feature vector generated at the bottleneck of the frontal generator (i.e., at the output of the encoder of the frontal U-Net generator) is passed through the decoder section of the profile U-Net generator to reconstruct the profile image. As we can see from  Fig. \ref{fig:profile_frontal_images}  and Fig. \ref{fig:frontal_profile_images},  the PF-cpGAN can preserve the identity and generate high-fidelity faces from an unconstrained dataset such as CMU-MultiPIE. These results show the robustness and effectiveness of PF-cpGAN for multiple use of profile to frontal matching in the latent common embedding subspace, as well as in the reconstruction of facial images.

\subsection{Evaluation  of the  Frontalization by CpGAN as a Pre-processing for Face Matching}\label{subsec:frontalization_vgg2}

As mentioned earlier, our coupled GAN framework can also be used for frontalization, which can be an important pre-processing step for other face-recognition tasks. Here, we conducted experiments to indicate the effectiveness of the frontalization performed using our cpGAN for the face verification task. In this set of experiments, we have used an Inception \cite{2015_GoogleNet} based FaceNet \cite{schroff2015facenet} model for the face verification task, which is specifically the NN2 model from \cite{schroff2015facenet}. We have performed this set of experiments on the VGGFace2 dataset \cite{VGGFace2}. 
 
 \begin{figure*}[t]
\centering     
\subfigure[Frontalization Performance]{\label{fig:front}\includegraphics[scale=0.44]{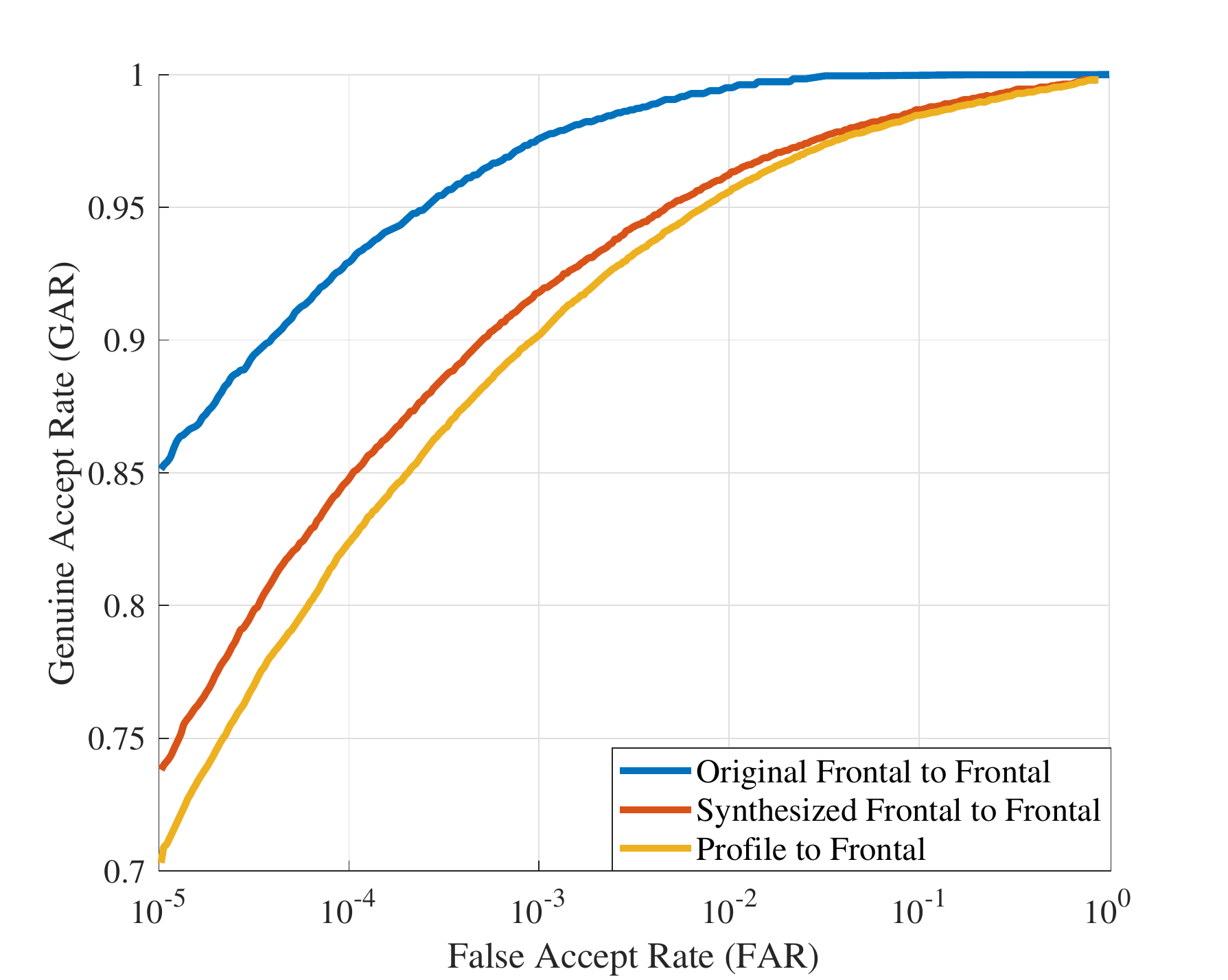}}
\subfigure[Different network comparisons]{\label{fig:GAN_imp}\includegraphics[scale=0.62]{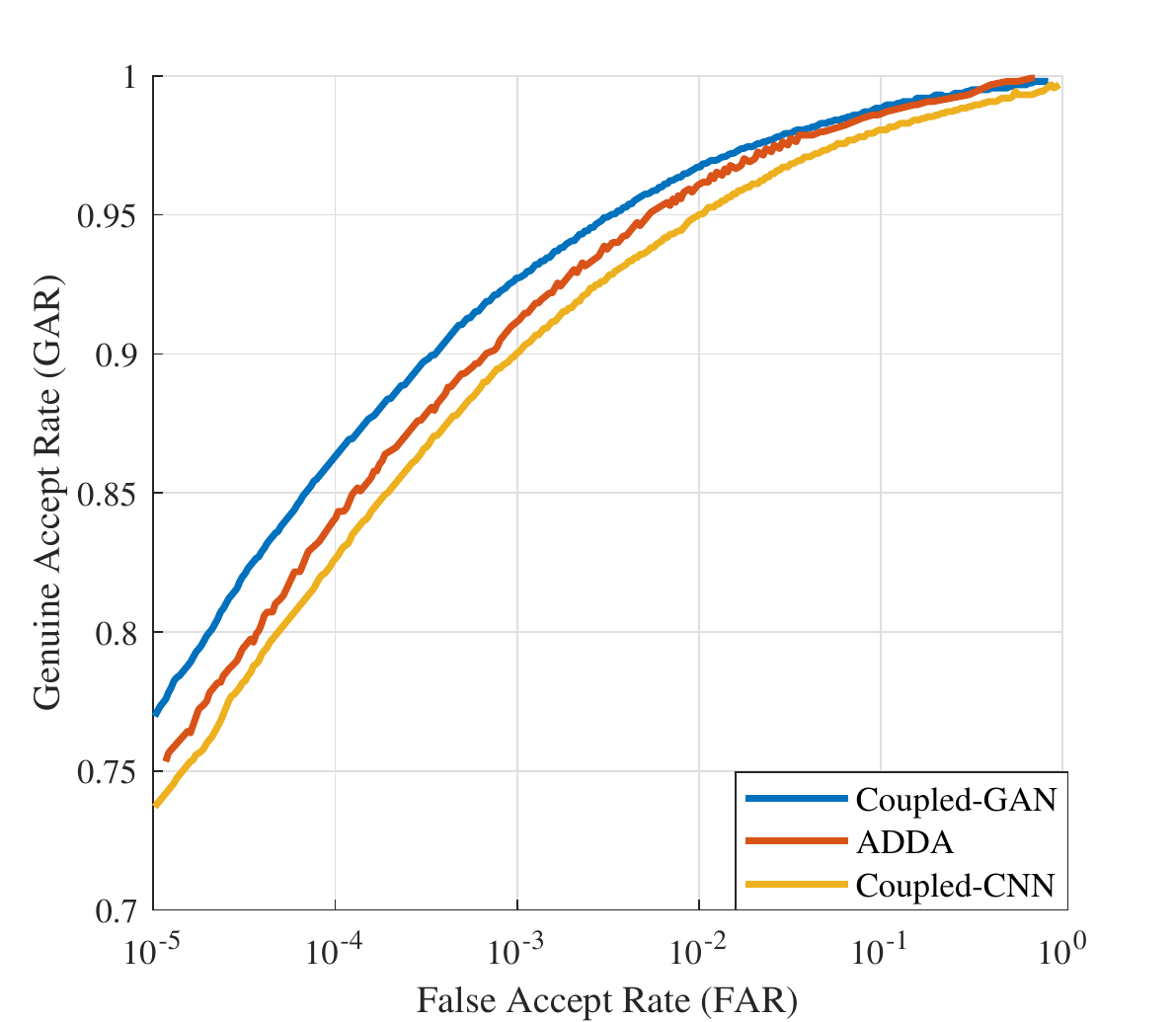}}
\caption{Performance comparison for (a) Frontalization using cpGAN as a preprocessing. (b) PF-cpGAN (Coupled-GAN) vs cpCNN (Coupled-CNN) vs PF-ADDA (ADDA). }
\label{fig:ROC_comp}
\end{figure*}

The VGGFace2 dataset provides annotation to enable evaluation of face matching across different poses \cite{VGGFace2}. In the dataset, six pose templates corresponding to three poses (i.e., two templates for a single pose) have been provided for about 300 identities. A template corresponds to five faces from the same subject with a consistent pose. This pose can be frontal, three-quarter or profile view. Consequently, for the 300 identities, there are a total of 1.8K templates with 9K images in total \cite{VGGFace2}. For this set of experiments, we have used only the profile and the frontal templates, which corresponds to about 6K images corresponding to 300 identities.

Here, we perform face verification using FaceNet in three different settings. In the first setting, we choose about 2.5K frontal images corresponding to 250 identities. Using these images, we fine-tune the Inception model NN2 from FaceNet for frontal to frontal face verification. Next, using this FaceNet model, we evaluate the frontal to frontal face verification on the remaining 50 identities. This setting will be called \emph{Original Frontal to Frontal}. In the second setting, we choose about 5K images corresponding to 250 identities, which have both profile and frontal images. Using these images, we fine-tune the Inception model NN2 from FaceNet for profile to frontal face verification. Next, using this FaceNet model, we evaluate the profile to frontal face verification on the remaining 50 identities. This setting will be called \emph{Profile to Frontal}. In the third setting, we used our cpGAN to frontalize the profile images from the dataset used in second setting (300 subjects with about 3K profile images) using the method outlined in Sec. \ref{subsec:recon}. We call this frontalized dataset synthesized frontal dataset. Next, using the fine-tuned FaceNet model from the first setting, we evaluate the frontal to frontal face verification on 50 identities from the synthesized frontal dataset. Specifically, in the third setting we are trying to check how well the proposed cpGAN is able to frontalize the images by running the frontal to frontal face verification model on the synthesized frontal dataset. This setting will be called  \emph{Synthesized Frontal to Frontal}. Note that we try to keep the 50 identities used for evaluation consistent across all the three settings.


Using the ROC curve as our performance metric, we have compared the performance of these three settings to evaluate the effectiveness of frontalization performed using our proposed cpGAN. The performance curves are provided in Fig. \ref{fig:front}. As expected the first setting (Original Frontal to Frontal) gives us the best performance and it is the upper bound as we are using the original frontal dataset for training and evaluation in this setting. On comparing the curves for the second (Profile to Frontal) and third settings (Synthesized Frontal to Frontal), it can be observed that the Synthesized Frontal to Frontal outperforms the Profile to Frontal face recognition model. This shows that the preprocessing in the form of frontalization performed using the proposed cpGAN framework improves the performance of a FaceNet model for profile to frontal face verification.

\subsection{Implementation of Coupled CNN and Domain Adaptation Network for Profile to Frontal Face Matching}
Before  the  advent of GAN,  many  deep-learning applications used CNNs for classification, regression, or reconstruction. To showcase the advantage of using a GAN model in our proposed approach for profile to frontal face recognition, we have also implemented two other frameworks that will be explained in this section. The performance comparison of the proposed PF-cpGAN with these new frameworks will be discussed in the following section.

\begin{figure}[t]
\centering
\includegraphics[width=5.0cm]{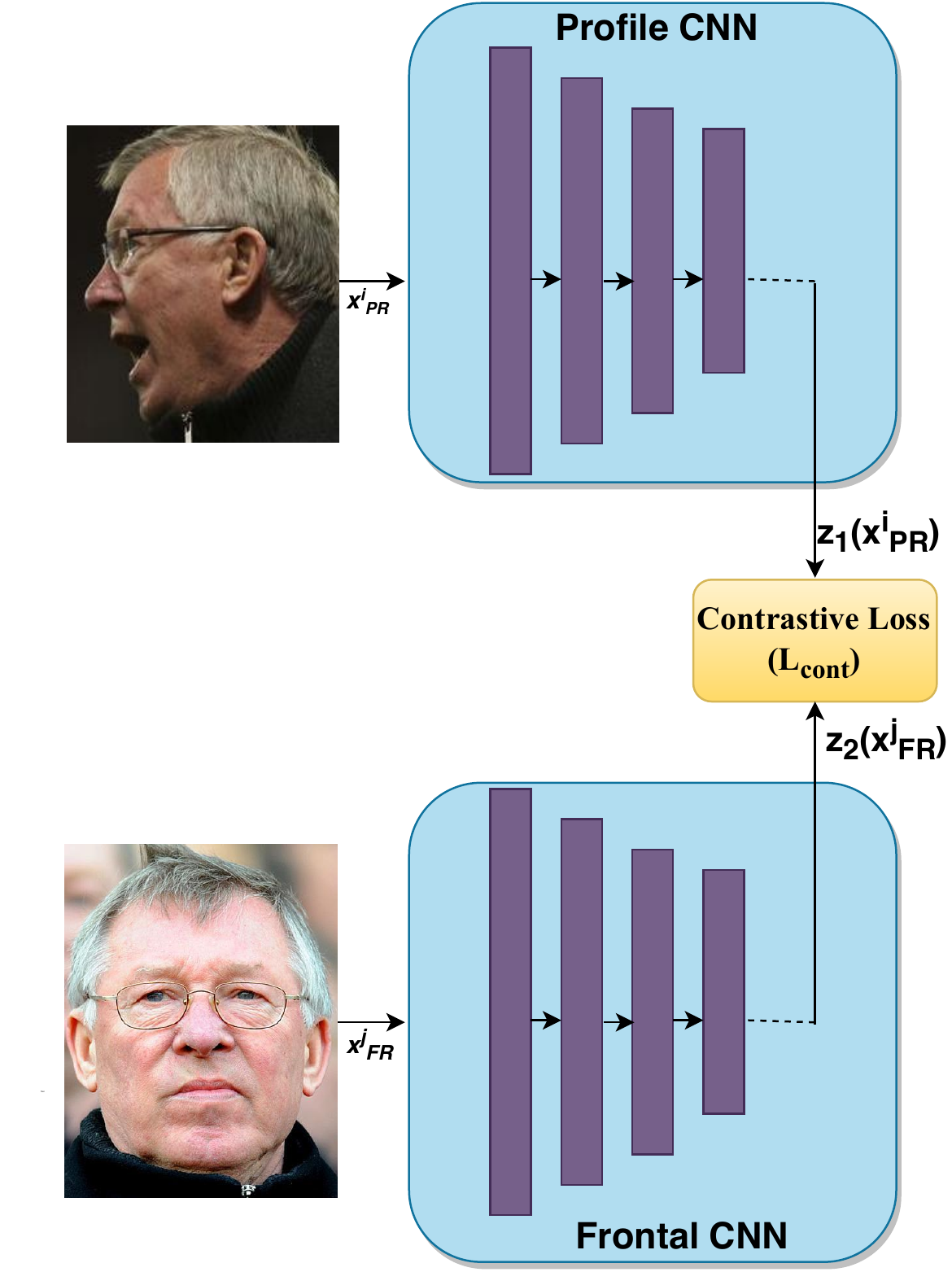}
\caption{Block diagram of Coupled CNN.}\label{fig:cnn_arch}
\end{figure}

    \textbf{Coupled CNN}: In the literature, it has been shown that GAN is better than CNN for some deep-learning applications. To confirm this hypothesis for our proposed application, we have implemented a coupled CNN and compared its performance with our proposed coupled GAN architecture. The coupled CNN (cpCNN) architecture is shown in Fig. \ref{fig:cnn_arch}. For fair comparison, we have used ResNet18 \cite{He2015_Resnet} pre-trained on ImageNet network as our CNN architecture for both Frontal CNN and Profile CNN. Additionally, we have added an extra fully-connected layer after the average pooling layer of ResNet18 for our coupled CNNs.
    
  The frontal and profile CNNs are coupled together at their output layer using a contrastive loss function $(L_{cont})$. This loss function  $(L_{cont})$ is a distance-based loss function, which is similar to the contrastive loss function (\ref{eq:5}) that we have used for PF-cpGAN. For ease of understanding, we have used the same naming convention for cpCNN as in  PF-cpGAN.
    
    
  
  We have used the VGGFace2 dataset for training and testing of the cpGAN. As in Sec. \ref{subsec:frontalization_vgg2}, we choose about 5K images corresponding to 250 identities, which have both profile and frontal images for fine-tuning the cpCNN for profile to frontal face verification. We have tested the cpCNN on the 50 disjoint identities from VGGFace2. The performance comparison is discussed in the following section.
 
 \begin{figure*}[h]
\centering
\includegraphics[width=15.5cm]{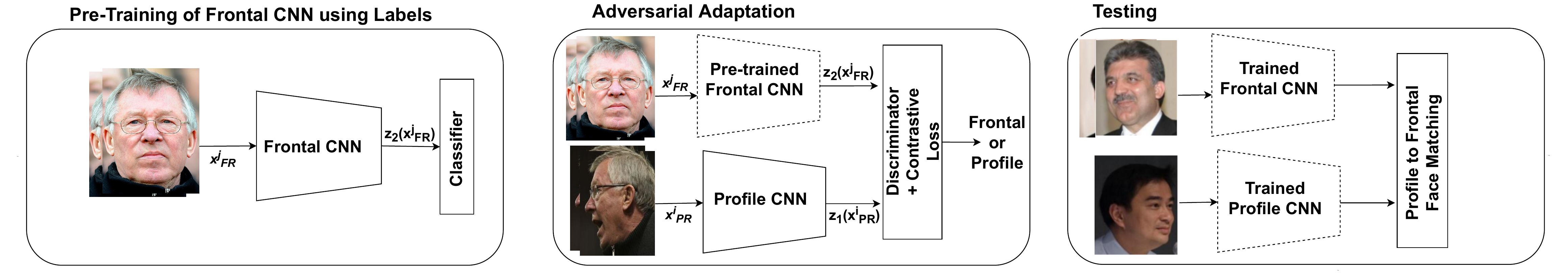}
\caption{Block diagram of Profile to Frontal Adversarial Domain Adaptation (PF-ADDA).}\label{fig:adda_arch}
\end{figure*}

\begin{figure*}[h]
\centering
\includegraphics[scale=0.33]{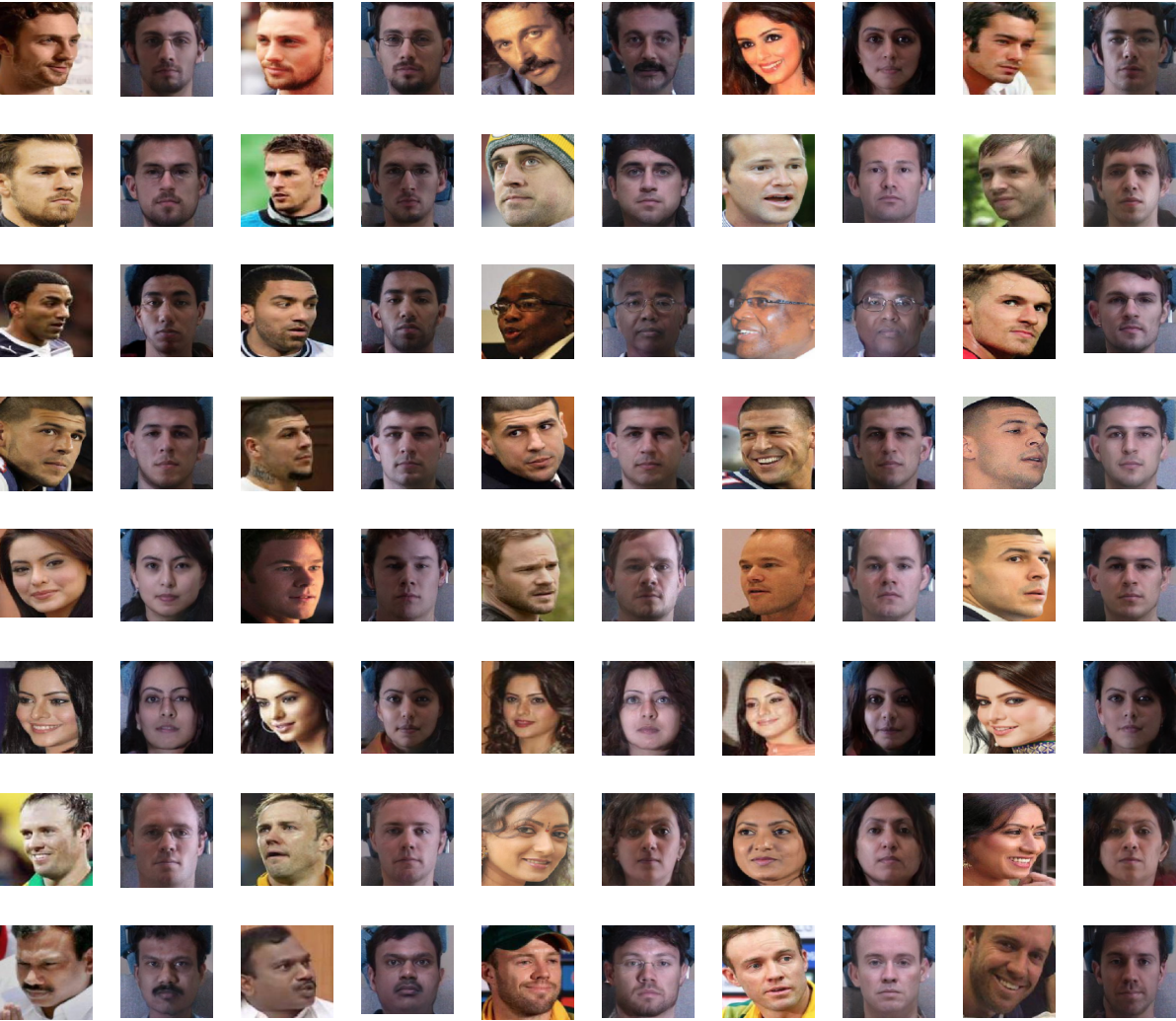}
\caption{Reconstruction of frontal images at the output of the  frontal U-Net generator with profile images as input to the profile U-Net generator. Every odd number column represents the input profile image, and every even number column represents the output frontal image. The input images belong to the VGGFace2 dataset.}\label{fig:frontal_profile_images1}
\end{figure*}

 \textbf{Domain Adaptation Network}: A profile to frontal recognition network could very well be implemented using deep-learning based domain adaptation techniques. These domain adaptation techniques attempt to alleviate the negative effects of domain shift (frontal domain to profile domain in our case) by learning deep neural transformations that map both domains into a common feature space. Recently, adversarial adaptation methods, which are based on reconstructing the target domain from the source representation have become increasingly popular. These adversarial methods seek to reduce an approximate domain discrepancy distance through an adversarial objective function with respect to a domain discriminator \cite{ADDA}.

 Taking a cue from \cite{ADDA}, we have implemented an unsupervised discriminative domain adaption network for profile to frontal face recognition. Hereafter, this network will be known as PF-ADDA. For this adversarial domain adaptation network, we consider the source domain as the frontal images and the target domain as the profile images. The architecture of PF-ADDA is shown in Fig. \ref{fig:adda_arch}. PF-ADDA has been implemented and optimized in two steps:
 \begin{itemize}
     \item In the first step of pre-training a frontal CNN, a discriminative representation is learned using the labels in the frontal image domain (source domain). This implies we first pre-train a frontal image encoder CNN using labeled frontal image examples. The optimization for this step is given as:
     \begin{equation}\begin{split}
     & \min_{z_2,C}  \; L_{cls}(X_{FR},Y_{FR}) = \\&- E_{(x_{FR},y_{FR})\sim (X_{FR},Y_{FR})}\sum_{k=1}^{K}\mathbbm{1}_{[k=y_{FR}]}\log C(z_2(x^j_{FR}))
     ,\end{split}\label{eq:20}  \end{equation} where the classification loss $L_{cls}$ is optimized over $z_2$, and frontal image classifier, C,  by training using the labeled source data, $X_{FR}, \text{and} \ Y_{FR}$.
 \item In the second step of adversarial adaptation, a separate encoding that maps the profile image data to the same space as the frontal image domain using an asymmetric mapping is learned through a combination of domain-adversarial loss and the contrastive loss. In other words, this implies that we perform adversarial adaptation by learning a profile image encoder CNN such that a discriminator that sees encoded frontal and profile images cannot accurately predict their domain label. In addition to the discriminator loss, the frontal and profile domain CNNs are also coupled through a contrastive loss.
The optimizations for this step are given as :  
 \begin{equation}\begin{split}
     \min_{D}  \; L_{adv_{D}}(X_{FR}&,X_{PR},z_2,z_1) = \\&- E_{x_{FR}\sim X_{FR}}[\log D(z_2(x^j_{FR}))] \\ & -
     E_{x_{PR}\sim X_{PR}}[\log (1-D(z_1(x^i_{PR})))]
     ,\end{split}\label{eq:21}
 \end{equation}
 \begin{equation}\begin{split}
     \min_{z_1}  \; L_{adv_{G}}(X_{FR}&,X_{PR},D) = \\&- 
     E_{x_{PR}\sim X_{PR}}[\log D(z_1(x^i_{PR}))]
     ,\end{split}\label{eq:22}
 \end{equation}and 
 \begin{equation}
\begin{split}
L_{cont}(z_1&(x^i_{PR}),z_2(x^j_{FR}),Y)= \\ & 
  (1-Y)\frac{1}{2}(D_z)^2 + (Y)\frac{1}{2}(\mbox{max}(0,m-D_z))^2.  
  \end{split}\label{eq:23}
  \end{equation}As shown in Equations (\ref{eq:21}), and (\ref{eq:22}), frontal image encoder CNN (source CNN) is fixed during the second stage, we just need to optimize the discriminator loss $L_{adv_{D}}$ and profile encoder loss $L_{adv_{G}}$ over the profile encoder CNN to generate $z_1$ without revisiting the  source domain encoder. Finally, along with the adversarial losses, we also optimize the contrastive loss,  $L_{cont}$, between the output of the Frontal CNN and Profile CNN as shown in (\ref{eq:23}). This contrastive loss is similar to the loss used for cpCNN and PF-cpGAN.\end{itemize}
     During testing, profile images (target domain) are mapped with the profile image encoder to the shared feature space, and  frontal images (source images) are mapped with the frontal image encoder to the shared feature space. Finally, the profile to frontal matching is performed in the shared feature space. Dashed lines in Fig. \ref{fig:adda_arch} indicate fixed network parameters.

  
  
We have used the VGGFace2 dataset for training and testing of the PF-ADDA. For fair comparison, the train and test split of the dataset for the PF-ADDA is consistent with the split for cpCNN.



\subsection{Performance Comparison of PF-cpGAN vs cpCNN vs PF-ADDA.} 
We have performed several experiments to compare the performance of our proposed  PF-cpGAN approach with cpCNN  and PF-ADDA. These experiments are performed on the VGGFace2 dataset \cite{VGGFace2}. As already mentioned in the previous section 5.7, for implementing cpCNN and PF-ADDA, we have used a common  network architecture as PF-cpGAN. Furthermore, we have been consistent in the training procedure (optimizer, batch-size, learning rate decay schedule, etc.). The performance comparison is plotted in terms of ROC and shown in Fig. \ref{fig:GAN_imp}. The  ROC results curves show that the proposed PF-cpGAN method outperforms other methods and gives much better performance for face verification under different pose variations. This demonstrates the effectiveness of coupled-GAN compared to other implementations. The improvement in performance using PF-cpGAN could be attributed to individual discriminators in the PF-cpGAN, which generate more domain specific features. The improvement can also be attributed to the sharpening of the features due to the perceptual loss terms.

\subsection{Coupled-GAN Qualitative Results on VGGFace2}
In this section, we test the robustness of our proposed approach Pf-cpGAN on VGGFace2 dataset by reconstructing frontal images from input profile images. In VGGFace2 \cite{VGGFace2}, two networks are trained to estimate the pose of images in the dataset. Specifically, a 5-way classification ResNet-50 is trained on the large-scale CASIA-WebFace dataset \cite{yi2014learning} to estimate head pose (roll, pitch, yaw). This model is then leveraged to predict pose of all the images in the dataset. As a result, VGGFace2 published different pose templates for 368 identities. Specifically, there are six templates for each subject: two templates each for frontal view, three-quarter view and profile view. There are five images per template. Here, we used about 250 identities to construct our pairs for training our coupled-GAN framework. Next, we test our network for frontalization of profile images. We follow the same procedure as discussed in Sec. \ref{subsec:recon} to frontalize our images. The results for frontalized images are shown in Fig. \ref{fig:frontal_profile_images1}. From these images, it can be observed that PF-cpGAN can preserve the identity and generate high-fidelity faces from VGGFace2 dataset. These results demonstrate the robustness and effectiveness of our coupled-GAN framework for frontalizing pose-variant images in the latent common embedding subspace.

\subsection{Ablation Study}\label{subsec:ablation}
The objective function defined in (\ref{eq:20}) contains multiple loss functions: coupling loss ($L_{cpl}$), perceptual loss ($L_{P}$), $L_2$ reconstruction loss ($L_2$), and GAN loss ($L_{GAN}$). It is important to understand the relative importance of different loss functions and the benefit of using them in our proposed method. For this experiment, we use different variations of PF-cpGAN and perform the evaluation using the IJB-A dataset. The variations are: 1) PF-cpGAN with only coupling loss and $L_2$ reconstruction loss ($L_{cpl} + L_2$); 2) PF-cpGAN with coupling loss, $L_2$ reconstruction loss, and GAN loss ($L_{cpl} + L_2 + L_{GAN}$); 3) PF-cpGAN with all the loss functions ($L_{cpl} + L_2 + L_{GAN} + L_{P}$).

\begin{figure}[t]
\centering
\includegraphics[scale=0.75]{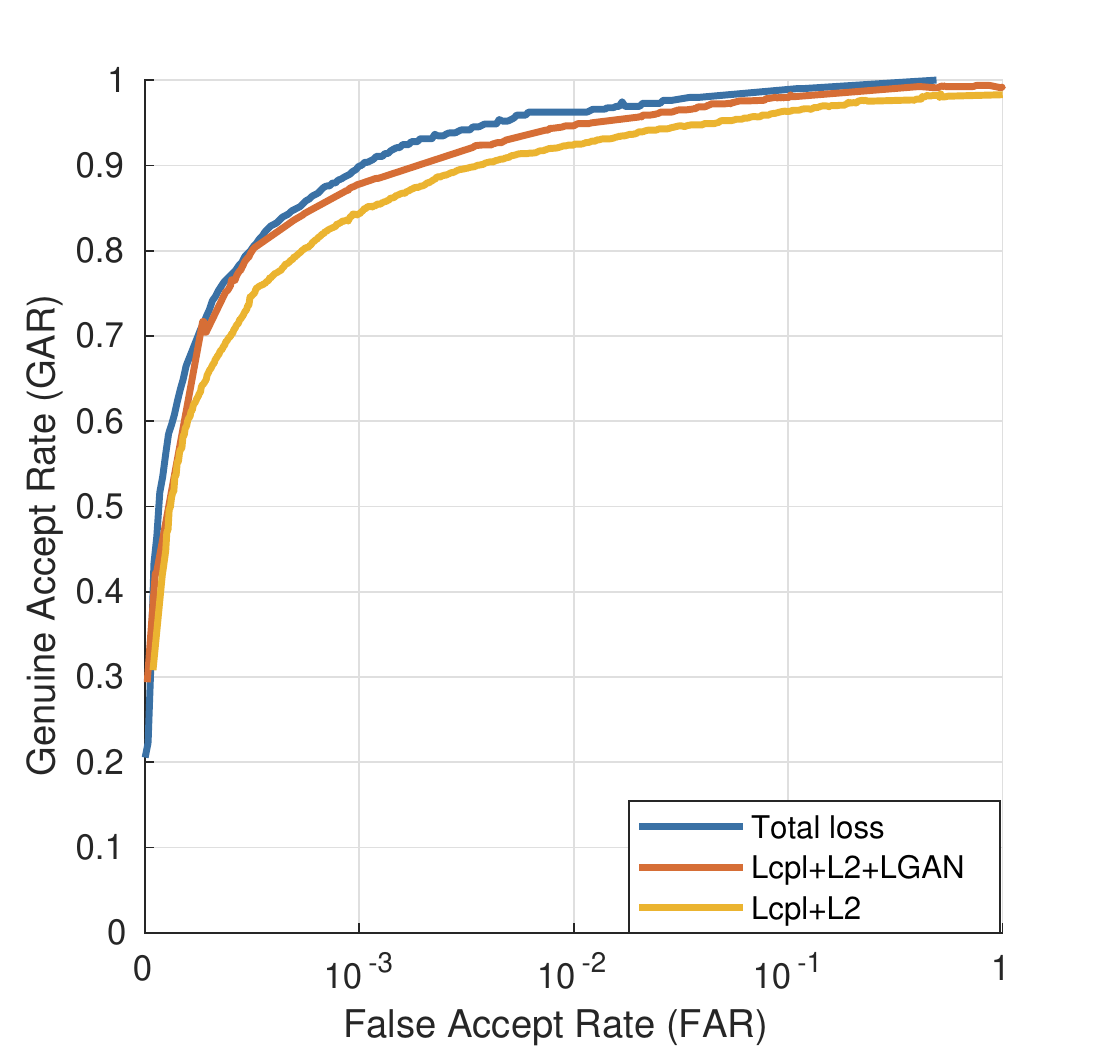}
\caption{ROC curves showing the importance of different loss functions for ablation study.}\label{fig:ablation}
\end{figure}


We use these three variations of our framework and plot the  ROC for profile to frontal face verification using the features from the common embedding subspace. We can see from Fig. \ref{fig:ablation} that the generative adversarial loss helps to improve the profile to frontal verification performance, and adding the perceptual loss (blue curve) results in an additional  performance improvement. The reason for this improvement is that using perceptual loss along with the contrastive loss leads to a more discriminative embedding subspace resulting in better face recognition performance.

\section{Conclusion}
   We  proposed a new framework which uses a coupled GAN  for profile to frontal face recognition. The coupled GAN contains two sub-networks which project the profile and frontal images into a common embedding subspace, where the goal of each sub-network is to maximize  the  pair-wise  correlation  between profile and frontal images during the process of projection. We  thoroughly evaluated our model on several standard datasets and the results demonstrate that our model notably outperforms other state-of-the-art algorithms for profile to frontal face verification. For instance, under extreme pose of $\pm90\degree$, PF-cpGAN achieves improvements of approx. $11\%$ (i.e., $77\%$ to $88\%$), when compared to the state-of-the-art methods for CMU-MultiPIE dataset. We have also explored two other similar implementations in the form of coupled CNN (cpCNN) and domain adaptation network (ADDA) for profile to frontal face recognition. We have compared the performance of the proposed approach with cpCNN, and ADDA and shown that the proposed approach performs much better than these two implementations. Moreover, we have also evaluated the frontal image reconstruction performance of the proposed approach. Finally, the improvement achieved by different losses including perceptual and GAN losses in our proposed algorithm has been investigated in an ablation study.

{\small
\bibliographystyle{IEEEtran}
\bibliography{submission_example}
}

\end{document}